\documentclass[a4paper,fleqn]{cas-dc}
\usepackage{pifont}
\usepackage[numbers]{natbib}
\usepackage{multirow}
\usepackage{graphicx}
\usepackage{float}
\usepackage{subcaption}

\usepackage{algorithm}
\usepackage{algorithmic}
\usepackage{hyperref}
\usepackage{xurl}        
\hypersetup{breaklinks=true}

\usepackage{svg}
\def\tsc#1{\csdef{#1}{\textsc{\lowercase{#1}}\xspace}}
\tsc{WGM}
\tsc{QE}
\tsc{EP}
\tsc{PMS}
\tsc{BEC}
\tsc{DE}

\begin{document}
\let\WriteBookmarks\relax
\def\floatpagepagefraction{1}
\def\textpagefraction{.001}

\shorttitle{}

\shortauthors{}

\title [mode = title]{\texttt\textbf{{LLM4EHR}}: Aligning Clinical Time Series with Medical Event Sequences via Large Language Models}                      

%
\author[1, 4]{Jingteng Li}[type=editor,
       orcid=0009-0003-0011-5828]
\corref{cor1}
\ead{ucabj23@ucl.ac.uk}

\credit{Writing - Original Draft, Writing - Review \& Editing, Data Curation, Visualization, Resources, Investigation, Formal analysis, Validation, Software, Methodology, Conceptualization}

\author[2, 3]{Alexander Capstick}[type=editor,
       orcid=0000-0002-3884-8045,]

\credit{Writing - Review \& Editing, Methodology}

\author[2, 4]{Louise Rigny}[type=editor,
       orcid=0009-0008-7252-6622,]

\credit{Writing - Review \& Editing}

\author[2, 4]{Iona Biggart}[type=editor,
       orcid=0009-0006-5785-4295,]

\credit{Writing - Review \& Editing, Methodology}

\author[1, 4]{Neil J Sebire}[type=editor,
       orcid=0000-0001-5348-9063,]

\credit{Writing - Review \& Editing, Supervision, Project administration, Funding acquisition}

\author[1, 2, 3, 4]{Payam Barnaghi}[type=editor,
       orcid=0000-0001-8591-9638,]
\corref{cor2}
\ead{p.barnaghi@imperial.ac.uk}

\credit{Writing - Review \& Editing, Methodology, Resources, Project administration, Supervision, Funding acquisition, Conceptualization}

\affiliation[1]{organization={GOS Institute of Child Health},
    addressline={University College London}, 
    city={London},
    country={United Kingdom}}

\affiliation[2]{organization={Department of Brain Sciences},
    addressline={Imperial College London}, 
    city={London},
    country={United Kingdom}}

\affiliation[3]{organization={UK Dementia Research Institute},
    addressline={Care Research and Technology Centre}, 
    city={London},
    country={United Kingdom}}

\affiliation[4]{organization={NIHR Great Ormond Street Hospital Biomedical Research},
    addressline={Centre and GOSH Data Research, Innovation and Virtual Environments}, 
    city={London},
    country={United Kingdom}}

\cortext[cor1]{Corresponding author}
\cortext[cor2]{Corresponding author}

\begin{abstract}
Recent research in clinical machine learning, focusing on outcome predictions in intensive care unit (ICU), has shifted from bespoke supervised models to foundation models, utilising modern representation learning methods. Here, foundation models are pre-trained on mixtures of complex clinical data modalities, useful for various downstream tasks. Existing works often utilise Electronic Health Records (EHR) to provide rich and diverse patient observations to train clinical foundation models. However, existing methods do not sufficiently explore the shared temporal structures between clinical events and time series (TS) observations recorded in EHRs. This limitation potentially leads to less robust and adaptive clinical foundation models, resulting in reduced performance on downstream tasks. To fully exploit this temporal structure, we propose LLM4EHR, a new clinical foundation model trained on ICU EHR data. Combining domain adapted large language models with a transformer TS encoder, we pre-trained LLM4EHR by temporally aligning the EHR events and TS. For this, we propose a regularised contrastive objective to learn robust EHR TS representations conditioned on EHR event embeddings produced by the domain adapted LLM. Supported by an ablation study, we find that learnt EHR TS embeddings from LLM4EHR improve performance on various downstream clinical tasks with competitive performance. Further, we empirically demonstrate that LLM4EHR learns transferable clinical TS embeddings that can be deployed to new cohorts via k-shot adaptation. These findings provide a step towards building more generalisable and performant clinical foundation models.
\end{abstract}


\begin{highlights}
\item LLM4EHR: Introduces a multimodal fusion framework that temporally aligns LLM‑encoded Electronic Health Record (EHR) event sequences with a dedicated EHR Time Series (TS) encoder via a semantic regularised contrastive objective.

\item Improved prediction outcomes: LLM4EHR achieved consistent performance gains on clinically relevant downstream tasks including mortality, phenotyping, decompensation, and remaining Length of Stay (LoS) prediction when compared to supervised and self-supervised baselines.

\item Improved transferability: We empirically demonstrate that LLM4EHR learns transferable clinical TS embeddings that can be deployed to new cohorts via k-shot adaptation.

\item Ablation and sensitivity: Ablation study involving different pre-training objectives showed that our semantic regularised alignment objective enabled LLM4EHR to achieve improved prediction outcomes in downstream tasks.

\end{highlights}


\maketitle

\section{Introduction}
\label{Sec_1}
\begin{figure*}[H]
\centering
\includegraphics[width=.9\linewidth]{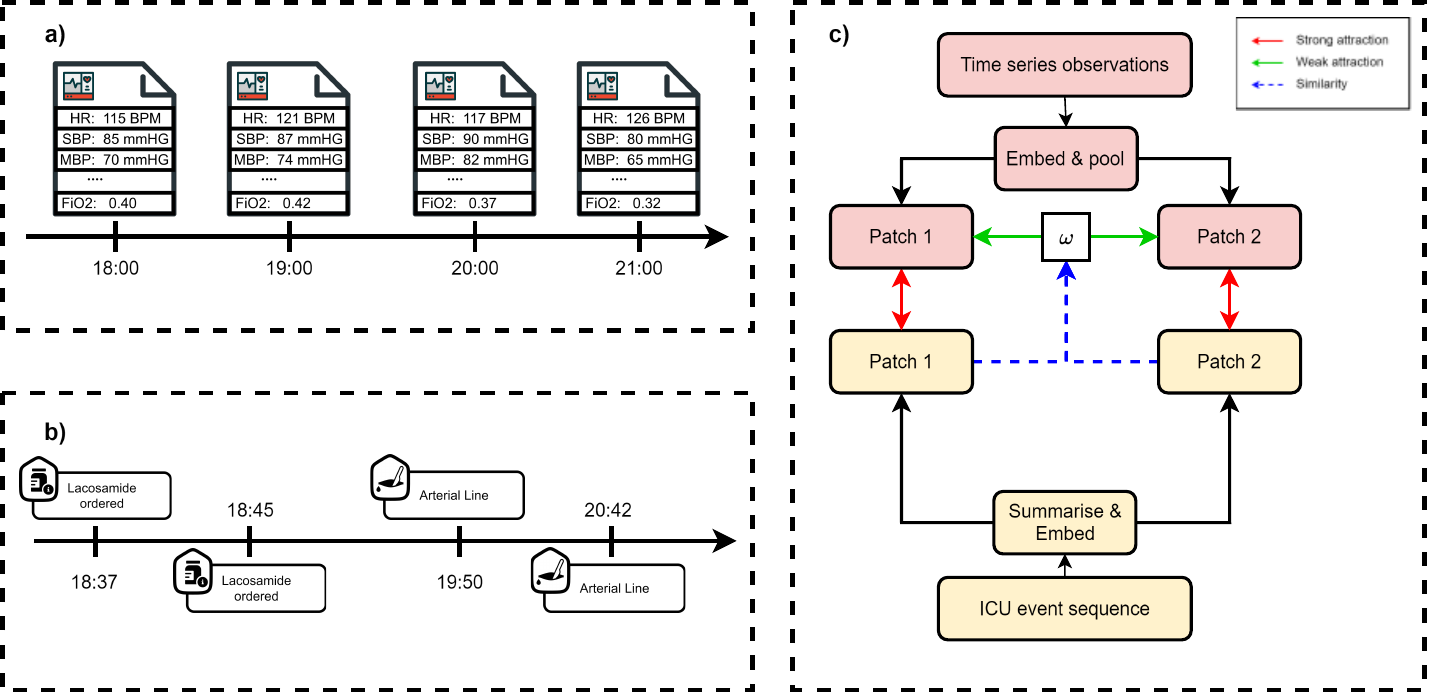}
\caption{a) An example of EHR TS, b) an example of EHR events and c) the difference between our regularised contrastive objective and the InfoNCE \cite{InfoNCE, he2020momentum} objective, during LLM4EHR pre-training. EHR events embeddings produced by a domain adapted LLM areused to guide the EHR TS embeddings via our proposed objective.}
\label{fig_1}
\end{figure*}

Electronic Health Records (EHRs) data is inherently multimodal. Temporal patient observations, interactions and measurements are recorded as EHR events and EHR Time Series (TS), as shown in \textbf{Figure \ref{fig_1} a} and \textbf{Figure \ref{fig_1} b}. Specifically, EHR events consist of itemised, irregularly sampled clinical interactions with specific time stamps, while EHR TS includes routinely collected patient measurements for patient monitoring. In EHR data modelling, entries such as medication prescriptions are typically treated as events due to the lack of fine-grained temporal dynamics. Conversely, monitoring signals such as patient vital signs are best modelled as TS due to the clinical significance of temporal variations in these signals. These two data modalities complement each other in representing a patient's health. However, the structural differences between these modalities pose a significant challenge for building versatile clinical foundation models. 

Towards overcoming this, we release a new structure for clinical foundation models, LLM4EHR, which jointly models EHR events and EHR TS via a time and semantic-aware contrastive pre-training objective. We find that the resulting learnt representations can adapt to downstream clinical tasks such as mortality prediction, phenotyping and remaining Length of Stay (LoS) prediction, which we support with an ablation study. Additionally, we show that LLM4EHR can be adapted to new datasets using routinely collected EHR TS variables with consistent performance.  

Traditionally, bespoke supervised methods were developed individually for clinical tasks, including survival analysis \cite{Ghassemi2014, Barajas2015}, risk prediction \cite{Alvarez2013, Cheng2016} and phenotyping \cite{Albers2014, Liu2015}. However, experiments conducted on large multitask clinical benchmarks \cite{MIMICBenchmark, eICUBenchmark} demonstrated that learnt information from one task is often relevant for others. Furthermore, patient observations in the clinical setting form temporal sequences, making them well-suited for language modelling techniques. These observations, along with recent findings that Large Language Models (LLMs) function as unsupervised multitask learners \cite{GPT-2}, have driven efforts to develop multitask clinical foundation models \cite{Wornow2023}. Recent work has thus explored using pre-trained LLMs for modelling EHR event trajectories \cite{BEHRT,MedBERT,ETHOS,Foresight} or EHR TS \cite{LLM4TS,AutoTimes}. A review by \citet{Wornow2023} found that clinical foundation models, pre-trained on large unlabelled EHR datasets, can adapt to downstream tasks using limited labelled data. 

However, many current methods fail to incorporate multiple temporal EHR data modalities during pre-training, despite evidence that combining modalities such as clinical notes and TS improves downstream performance \cite{king2023,MedsTsLLM,MAIN}. Given the interdependency between EHR events and EHR TS observations, we hypothesise that similar improvements could be achieved by jointly modelling these two temporal modalities via self-supervised pre-training. 

Prior work aligning these different clinical data modalities focused on contrastive representation learning between pairs of samples \cite{Kline2022}. In these cases, the alignment is often centred on shared identities between modalities, such as aligning the clinical TS and note embeddings describing the same care period \cite{king2023,MAIN}. However, this stationary, instance-wise alignment across clinical data modalities does not consider the shared temporal structure between EHR events and EHR TS observations, which we hypothesise provide more effective learnt representations.

Unlike prior work, LLM4EHR learns entangled EHR TS representations with EHR events embeddings, where a trainable EHR TS encoder is guided by a pre-trained and domain-adapted large language model (LLM) via contrastive pre-training. We reformulate the commonly used InfoNCE \cite{InfoNCE,he2020momentum} objective to temporally align the EHR TS embedding with EHR event embedding produced by a domain-adapted LLM, where EHR TS embedding is pulled closer to EHR event embedding at each time step. This leads to improved performance in downstream tasks that generalise to new datasets.

Additionally, we proposed incorporating learnt semantic similarities from the pre-trained LLM into the contrastive objective to mitigate the effects of class collision \cite{SimCLR, Zheng2021, Wu2024} in contrastive learning. Class collision arises when similar instances are pushed apart in the embedding space due to strict one-to-one positive sampling. If EHR TS observations at $t$ and $t'$ correspond to similar EHR events, they should not be treated as negatives, as implied by the InfoNCE loss. We present a semantic-regularised InfoNCE objective \cite{InfoNCE, he2020momentum}, where the semantic similarities between LLM-embedded EHR events are used to weigh the learnt EHR TS embeddings (\textbf{Figure \ref{fig_1} c}), providing improved EHR TS embeddings. Our contributions are as follows:

\begin{itemize}
    \item We present LLM4EHR, an LLM-based framework for representation learning of EHR TS, trained on data from intensive care units.
    
    \item To train LLM4EHR, we propose a contrastive learning objective to temporally align EHR events and EHR TS embeddings. With an ablation study, we find that this improves the learnt representations.
    
    \item We show that our method achieves superior performance in clinical classification tasks in few-shot and cross-dataset settings.
 
\end{itemize}

\begin{figure*}[!t]
\centering
\includegraphics[width=.8\linewidth]{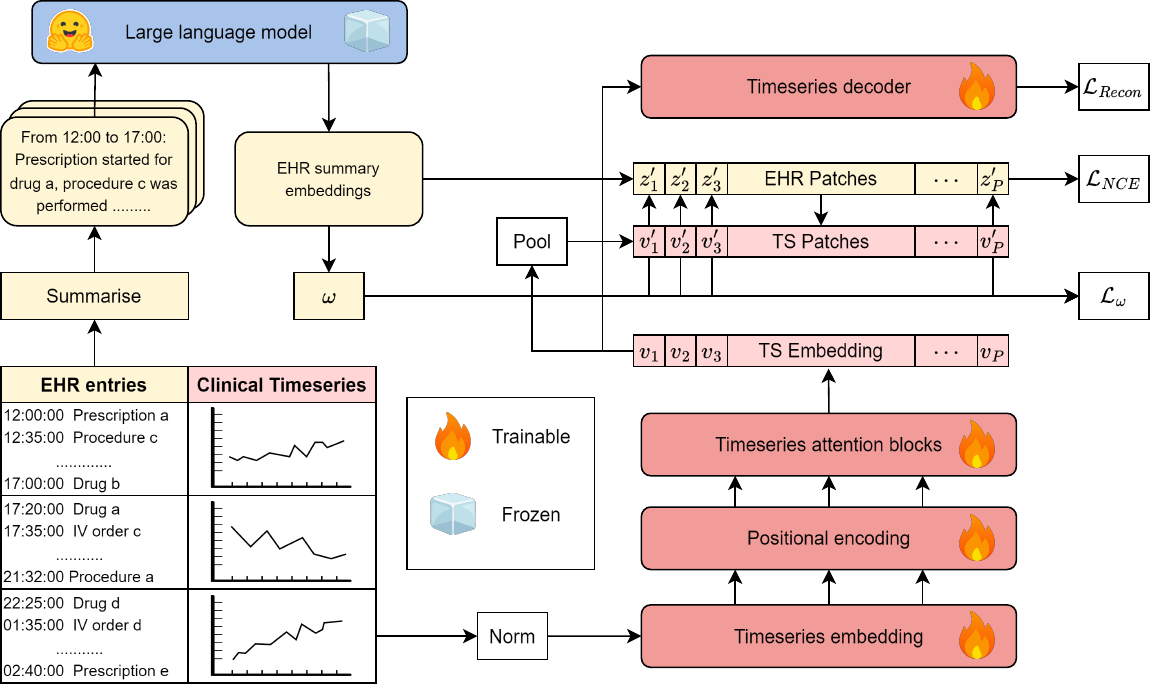}
\caption{Overview of LLM4EHR, EHR entries and clinical TS are centred on ICU admissions, where EHR operations are coloured in beige and TS operations are coloured in red, pre-trained LLM layers (besides the embedding weights for new tokens) are frozen during pre-training.}
\label{fig_2}
\end{figure*}

\section{Related work}
\label{Sec_2}
EHR foundation models are clinical foundation models trained on large-scale EHR data \cite{Wornow2023}. Recent methods exhibit notable similarities with LLMs in that they are trained on tokenised event sequences via autoregressive \cite{ETHOS,Foresight,MedGPT,Cascella2023} or masked sequence modelling \cite{MedBERT,BEHRT,BioClinicalBERT,CEHR-BERT}. Notable examples include Foresight by \citet{Foresight} and ETHOS by \citet{ETHOS} , which repurposed the autoregressive attention layers of GPT-2 \citep{GPT-2} to simulate event transitions in patient timelines. However, these works primarily treat patient trajectories as event sequences, which can lead to fine-grained temporal dynamics being lost or obscured during modelling.

Self-supervised representation learning for TS learns transferable representations of TS features from unlabelled data for downstream tasks \cite{SimMTM,PatchTST,PatchTSMixer}. Recent work demonstrated improved performance in TS forecasting by modelling TS via transformer architectures \cite{Transformer}. Here, TS segments are tokenised to form temporal sequences, and attention layers are trained to process TS tokens via self-supervised pretext tasks. LLM4TS by \citet{LLM4TS} used self-supervised reconstruction to repurpose pre-trained LLMs for TS forecasting. Similarly, AutoTimes by \citet{AutoTimes} used language prompts during self-supervised pre-training to allow in-context TS forecasting via frozen LLMs. Different from prior works, LLM4EHR maintains a dedicated transformer encoder for EHR TS, where EHR TS embeddings are guided by EHR event embeddings produced by a pre-trained LLM during pre-training.

Contrastive learning can create entangled representations between pairs of samples. Prior works in clinical machine learning explored aligning similar EHR TS samples via contrastive learning. EBCL by \citet{EBCL} used contrastive pre-training to align TS observations before and after key index events and showed improved performances in downstream classification tasks. CROCS by \citet{Kiyasseh2021} used contrastive learning to align ECG measurement embeddings with patient prototype embeddings, where patient prototypes serve as queries to retrieve similar ECG samples. \citet{king2023} explored aligning clinical time series (TS) with ICU notes via a shared LLM. However, their work maintains a stationary alignment between EHR TS and clinical notes and does not consider important temporal relationships.

\section{Background}
\label{Sec_3}
\subsection{EHR data representation}
\label{sec::backgroud_1}
When patients interact with care services, each measurement, observation, and care giver action is recorded in an Electronic Health Record (EHR). While different types of EHR data share the same structured format, they are generated through different processes and often use different data modalities. Regularly collected physiological observations are usually modelled as clinical Time Series (TS) due to their dynamic nature. These variables are tracked over time to monitor a patient's health status, with abrupt changes contributing heavily to standardised risk analyses such as SAPS \cite{SAPS} or SOFA \cite{SOFA}. Supervised methods in clinical machine learning often rely on TS features as input for downstream tasks, especially for time-sensitive predictions such as survival \cite{Ghassemi2014, Barajas2015} or risk \cite{Albers2014, Liu2015} predictions. 

In contrast, certain events or occurrences in EHR systems are not conveniently or logically modelled as TS variables. For example, while the presence of a standing drug prescription at a given time is clinically meaningful, modelling it as a TS variable is not informative due to the lack of temporal variation. Additionally, each unique event would be represented as a separate binary variable, which scales poorly given the natural diversity of unique interactions in EHR systems. EHR events are typically tokenised to form temporal sequences for sequence modelling \cite{Meds_reader, ETHOS, Foresight}, which simplifies the temporal dimension as longitudinal ordering of tokens. 

Similar to prior work on cross-modal alignment between EHR TS and clinical notes, we define the problem of modelling multi-modal EHR data as cross-modal alignment between EHR TS and events, where TS variables and EHR events are generated separately but aligned in time. We did not use clinical notes in our framework, as while clinical notes provide a high-level summary of patient observations and treatments in the ICU, they are often documented with substantial delay and therefore cannot be assumed to align temporally with fine-grained TS observations. 

We define clinical time series as multivariate time series consisting of time varying patient observations, primarily lab and nurse charted lab observations. While some work does not treat irregularly sampled lab values as TS variables \cite{stein2025prediction}, we included them as changes in lab observations over time are clinically significant. Formally, we define an instance of clinical TS with $L$ variables over $T$ time steps as $x^{1:T} = \{x^{1},x^{2}, \cdots, x^{T}\}, x^{t}\in \mathbb{R}^{L}$.

EHR events, however, are represented as time-indexed sequences. We define an EHR event sequence with $N$ events over $T$ as $k^{1:T} = \{k_{n_{1}}^{1},k_{n_{2}}^{2},\cdots,k_{n_{T}}^{T}\}$, where $k_{n_{t}}^{t}\in\mathcal{K}$ is a set of $n_{t}$ clinical events observed at time step $t$. In this case, we treat time-stamped care giver actions as EHR events. In MIMIC-IV, these entries include prescription orders, medication administrations, infusion orders, renal replacement therapies (RRT/CRRT) settings, oxygen-delivery settings and the insertion, presence or removal of invasive lines. Since these events are not regularly sampled or distributed evenly overtime, the number of events $n_{t}$ at time $t$ is not fixed.

Finally, we consider data collected for a given patient throughout a single Intensive Care Unit (ICU) stay as an 'episode'. An episode starts at ICU admission and ends at ICU discharge. A full episode is defined as $e_{i} = (x_{i}^{1:T_{i}}, k_{i}^{1:T_{i}})$, where both $x_{i}$ and $k_{i}$ are centred on ICU admission and are aligned with the time indicator $t\in T_{i}$.

\subsection{Multimodal pre-training of EHR data}
\label{sec::backgroud_2}
Pre-training on clinical TS aims to learn general temporal patterns in patient trajectories using a task agnostic objective. The goal of the pre-training objective is to reward the model for capturing clinically meaningful dynamics, such as trends or sudden changes in physiological observations, without bias toward any specific downstream task. Because EHR TS and EHR events are generated through separate processes but are temporally aligned, we hypothesise that explicitly aligning these two modalities during pre-training allows TS encoders to learn more informative EHR TS representations, leading to better performance in context sensitive downstream tasks such as decompensation and Length-of-stay predictions.

Existing multimodal pre-training frameworks for EHR data involve stationary alignment of two EHR modalities via a contrastive objective. Here, we focus on multimodal pre-training methods involving non-imaging data. The most common form of alignment occurs between clinical TS and clinical notes, where the sequence-wise representation of EHR TS across an episode is pulled closer to the embedding of clinical notes describing the same episode. The best example is the multimodal pre-training framework by \citet{king2023}, where a masked attention encoder is used to encode clinical TS and the resulting embedding is aligned with the clinical note embedding produced from a domain adapted BERT model \cite{BioClinicalBERT} via a contrastive objective. We argue that stationary alignment between clinical TS and notes is insufficient due to a mismatch in temporal granularity, as aligning fine-grained TS representations with coarse-grained clinical note embeddings fails to adequately capture the shared temporal dynamics between the two modalities. 

A more recent work by \citet{CTPD} explored learning shared temporal patterns between clinical TS and notes, where TS and clinical notes embeddings are first aligned at the attention level via a contrastive objective, expressed over individual time steps, and then concatenated at the final level for final output. We note that CTPD is susceptible to time mismatches between EHR TS and clinical notes, as clinical notes are not synchronised with EHR TS observations. Additionally, CTPD explicitly assumes that clinical notes are available during prediction, which is counterintuitive for time-sensitive predictions, as clinical notes are not produced at the point of observation and are unlikely to be available at the prediction time. 

Additionally, the semantic knowledge in pre-trained LLMs is not utilised in either framework. Clinical LLMs are used only as fixed encoders for clinical notes, without modifying the pre-training objective to transfer semantic structure into the TS representation. Therefore, we propose a similar framework that aligns EHR event summaries with EHR TS embeddings through an explicit temporal alignment objective derived from InfoNCE, while also leveraging the LLM's learnt semantic similarity to guide the EHR TS representation.

\section{Methods}
\label{Sec::method}

\subsection{LLM4EHR structure}
\label{Sec::method_1}

\begin{table}[h]
\caption{Example aggregated EHR summary for a 5‑hour window (70–75 hours after ICU admission), the patient was given Furosemide, pantoprazole by the care givers and received scheduled Insulin infusion, no invasive lines and renal replacement therapies were active.}
\label{tab_ehr}
\resizebox{\columnwidth}{!}{%
\begin{tabular}{ll}
\hline
Summary & Text \\ \hline
Prescriptions & \begin{tabular}[c]{@{}l@{}}Prescription Orders:\\ Prescriptions started: \\ Furosemide (40mg/4ml Vial), \\ Pantoprazole (40mg Tablet)\\ Prescriptions ended: \\ Furosemide (40mg/4ml Vial)\end{tabular} \\ \hline
emar & \begin{tabular}[c]{@{}l@{}}Administered:\\ Furosemide, \\ Insulin, \\ Pantoprazole, \\ Remdesivir\end{tabular} \\ \hline
Infusion & \begin{tabular}[c]{@{}l@{}}Infusion Order:\\ - NaCl 0.9\% 250ml (active)\\ - Furosemide (Lasix) 40.00mg (active)\\ - Insulin - Humalog 4.00 units (active)\end{tabular} \\ \hline
rrt/crrt & No active renal replacement therapy \\ \hline
Oxygen & \begin{tabular}[c]{@{}l@{}}Delivery device: \\ Oxymizer, High flow nasal cannula\\ \\ Oxygen flow: \\ 6.0\\ \\ Additional Oxygen flow: \\ None\end{tabular} \\ \hline
Invasive lines & No invasive lines active \\ \hline
TS & \begin{tabular}[c]{@{}l@{}}- albumin: \\ mean: 3.50; min: 2.80; max: 3.33; std: 0.28; count: 2\\ - no measurements for fio2\\ - no measurements for Tropinin T\\ ......\end{tabular} \\ \hline
\end{tabular}%
}
\end{table}

EHR events and EHR TS provide complementary views of a patient's ICU stay. We hypothesise that conditioning TS representations on information from EHR events yields more informative and generalisable TS embeddings for downstream clinical tasks, such as predicting intensive care outcome or patient decompensation. Given a patient's full EHR episode $e_{i}$, consisting of EHR event sequence $k_{i}^{1:T}$ and multivariate EHR TS $x_{i}^{1:T}$, LLM4EHR pre-training aligns the EHR events embedding $z_{i}^{1:T}$ with the TS embeddings $v_{i}^{1:T}$ via a contrastive objective applied at each time step and aggregated over $T$.

Aligning EHR events and TS embeddings directly step-by-step is problematic due to the structural differences between EHR events and EHR TS data. First, as explained in \textbf{Section \ref{sec::backgroud_1}}, EHR events are not regularly sampled, and the number of events $n_{t}$ at time step $t$ is not fixed. Expressing a contrastive objective directly over $T$ results in an ambiguous objective. Additionally, jointly learning EHR TS and EHR event embeddings is unstable for contrastive learning. Since the learnt EHR TS embedding is conditioned on EHR events embedding in LLM4EHR, rapidly changing the EHR events embedding reduces the consistency of the alignment objective \cite{he2020momentum}.

We propose using a domain adapted LLM to embed EHR events, similar to prior work by \citet{Sun2024}. We define the EHR event set $\mathcal{K}$ as a set of natural language phrases describing each unique EHR event, such as the name and dosage of a prescription order. Different from \citet{Sun2024}, multiple EHR events at $t$ are aggregated into a single natural language summary using a fixed schema. An example of the aggregated summary is shown in \textbf{Table \ref{tab_ehr}}. Here, $z_{i}^{1:T}$ is produced by a frozen LLM domain adapted using clinical domain texts and $v_{i}^{1:T}$ is produced by a trainable TS encoder, as shown in \textbf{Figure \ref{fig_2}}.

In MIMIC-IV, we found that aligning EHR summary embeddings with EHR TS embeddings hourly is suboptimal due to the relative sparsity of EHR events, and therefore enforced a bin width of $5$ hours. In practice, a small bin would result in many instances of $v_{i}^t$ with no corresponding $z_{i}^t$. Conversely, a large bin would degrade the temporal consistency of the alignment process, similar to the temporal mismatch between EHR TS and note alignment in \textbf{Section \ref{sec::backgroud_2}}. EHR TS variables within each time step are aggregated with windowed features (mean, min, max, std and count). Finally, we append a plain text summary of TS variables to the end of EHR event summary at $t$ to provide an additional layer of mutual information between the two modalities.

\subsection{Semantic alignment}
\label{Sec::method_2}

\begin{figure}[]
  \centering
  \includegraphics[width=\columnwidth]{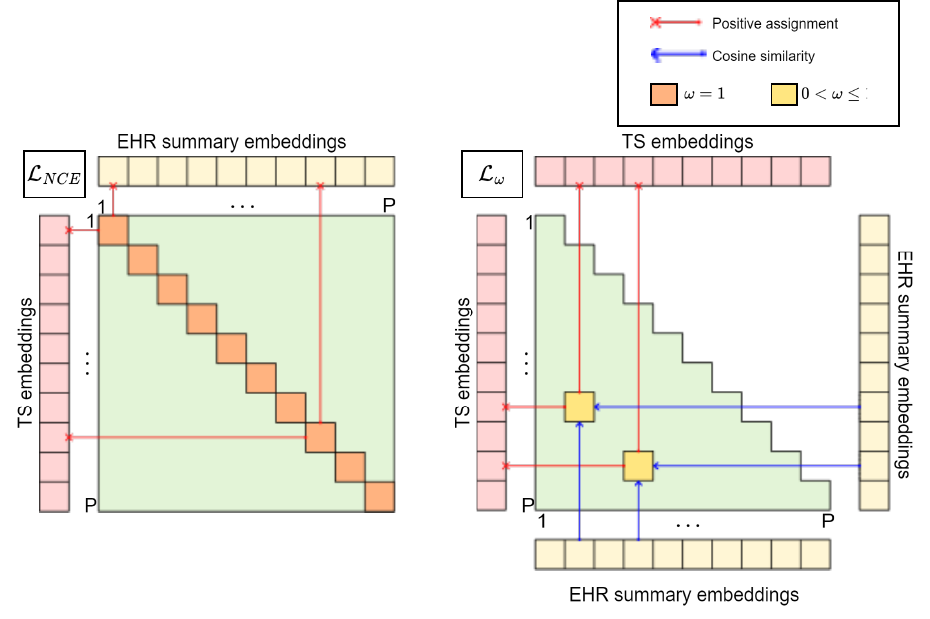}
  \caption{ Calculating the semantic alignment loss for a given episode; the figures only include the lower triangular and non-diagonal part of the soft assignment $\omega$ since $\omega$ is symmetric.}
  \label{fig_omega}
\end{figure}

The attraction between elements of $v_{i}^{1:T}$ and $z_{i}^{1:T}$ is defined at each timestamp via a modified InforNCE \cite{InfoNCE} loss. Similar to \citet{he2020momentum} and \citet{Kiyasseh2021},  we implemented the normalised variant of InfoNCE $\mathcal{L}_{\text{NCE}}$, with temperature $\tau$ as:

\begin{align}
\label{eq_2}
    &\mathcal{L}_{\text{NCE}} = -\sum_{t=1}^{T}\log \frac{\exp{(\text{sim}(v^{t}, z^{t})/\tau)}}{\sum_{t'\in T\backslash t}\exp{(\text{sim}(v^{t},z^{t'})/\tau)}} \\ &\text{sim}(v^{t},z^{t}) = \frac{v^{t}\cdot z^{t}}{\lVert v^{t} \rVert \lVert z^{t} \rVert}
\end{align}

However, we note that this objective is susceptible to class collision, where instances within same class are pushed apart due to the loss being defined between an instance and its augmented view. In our case, similar instances of TS observations should remain closer after embedding. Since LLMs are known to capture word similarities \cite{Mahajan2025}, we propose a $\omega$-regularised variant of the InfoNCE loss to weight similarities between instances within $v_{i}^{1:T}$ based on the semantic similarities between instances within $z_{i}^{1:T}$. Following \textbf{Equation \ref{eq_2}}, the regularisation function $\mathcal{L}_{\omega}$ is given as: 

\begin{align}
    \label{eq_3}
    &\mathcal{L}_{\omega} = -\sum_{t=1}^{T}\sum_{\substack{t^{*}=1 \ t^{*}<t}}^{T}\omega_{t,t^{*}}\log\frac{\exp(\text{sim}(v^{t},v^{t^{*}})/\tau)}{\sum_{t'\in T\backslash t}\exp{(\text{sim}(v^{t}, v^{t'})/\tau})} \\ &\omega_{t,t^{*}} = \frac{\exp(\text{sim}(z^{t}, z^{t^{*}}))}{\sum_{t'\in T\backslash t}\exp(\text{sim}(z^{t}, z^{t'}))}
\end{align}

The attraction weight $\omega_{t,t^{*}}$ is interpreted as the degree of agreement between $v^{t}$ and $v^{t^{*}}$, quantified using the similarity between $z^{t}$ and $z^{t^{*}}$. For a pair of $v^{t}$ and $z^{t}$, the weight $\omega$ is a $T\times T$ symmetric matrix, as shown in \textbf{Figure \ref{fig_omega}}. We thus used only the lower triangular part of $\omega$, such that $\omega_{t,t^{*}}$ is only defined for $t^{*}<t$. Intuitively, \textbf{Equation \ref{eq_3}} pulls TS observations corresponding to similar EHR events closer, such that similar instances within $x_{i}$ are consistent after embedding. The overall alignment objective is defined as:

\begin{equation}
\label{eq_4}
    \mathcal{L}_{\text{align}} = \mathcal{L}_{\text{NCE}}+\mathcal{L}_{\omega}
\end{equation}
    
\subsubsection*{Pre-training}
\label{Sec::method_3}

The EHR summary embedding $z_{i}^{t}$ is obtained by passing the EHR summaries for episode $e_{i}$ at $t$ through the pre-trained LLM, then applying masked average pooling over the hidden states after the last attention block. Since EHR summaries contain complex clinical terminologies such as specific drug compounds or ventilator settings, we used domain adapted LLMs to embed EHR summaries. We selected the BioClinical ModernBERT \cite{BioClinicalModernBert}, a version of ModernBERT \cite{ModernBERT} further trained on clinical language datasets and with a long context length of 8192. Because gradients cannot propagate through the frozen LLM, an additional dense projection layer is added for training stability. The final EHR summary embedding $z_{i}^{t}$ is produced by passing the average-pooled EHR summary embedding through the projection layer. During pre-training, we update the weights of the projection layer with a large momentum ($m=0.9$) via the Adam optimiser \cite{kingma2014adam} to keep $z_{i}^{t}$ stable for calculating $\mathcal{L}_{\text{align}}$.

As shown in \textbf{Figure \ref{fig_2}}, the EHR TS encoder is a separate, fully-trainable encoder for encoding EHR TS. We built the EHR TS encoder with 1D convolutional embedding layers, trainable positional embeddings for positional encoding, and autoregressive attention blocks \cite{GPT-2}. Similar to LLM4TS by \citet{LLM4TS} and AutoTimes by \citet{AutoTimes}, we treat instances of $x_{i}^{t}$ as tokens and employ autoregressive attention blocks to model the transition between TS observations.

The EHR TS embedding $v_{i}^{1:T}$ is produced by passing z-standardised EHR TS features through the embedding layers, with added positional embeddings from passing the EHR TS positional encodings through the positional embeddings layers. The EHR TS embeddings are then passed through $N$ autoregressive attention blocks before calculating the alignment loss via \textbf{Equation \ref{eq_4}}. To produce our final optimisation objective, we reconstruct the input TS autoregressively from the latent representation $v_{i}$ and use the reconstruction loss as an auxiliary objective to $\mathcal{L}_{\text{align}}$. This provides the loss value:

\begin{equation}
\label{eq_5}
    \mathcal{L}_{\text{total}} =  \mathcal{L}_{\text{align}} + \mathcal{L}_{\text{recon}} = \mathcal{L}_{\text{align}}+ \frac{1}{T}\sum_{t=2}^{T}\rVert x_{i}^{t} - \hat{x_{i}}^{t}\lVert^{2}_{2}
\end{equation}

After pre-training, LLM4EHR can be adapted to downstream clinical tasks using task-specific linear heads, as with other TS foundation models. During linear probing, only EHR TS observations are used to make downstream predictions. The TS embedding layer, positional embedding layer and TS encoder attention blocks are kept frozen during fine-tuning. For complex longitudinal prediction tasks, such as rolling patient decompensation and length-of-stays predictions, we unfreeze the last TS encoder attention block to give LLM4EHR more flexibility to adapt to the unique temporal dynamics of these downstream tasks.

\section{Results}
\label{Sec_4}

\subsection{Datasets}
\label{result_1}

We evaluated the performance of our pre-trained model on two public intensive care EHR datasets: MIMIC-IV \cite{mimiciv} and the Physionet Challenge 2012 (Physionet2012) \cite{Physionet2012} (\textbf{Table \ref{tab_dataset}}). We conducted evaluations using four downstream ICU outcome prediction tasks in MIMIC-IV and used Physionet2012 as an external dataset to assess the transferability of models pre-trained on MIMIC-IV. For MIMIC-IV, we retained the first ICU stay for each unique hospital admission and excluded stays shorter than 48 hours. Each ICU episode includes 32 physiological and laboratory variables as EHR TS variables, derived from nursing and laboratory observations recorded during the episode. These variables are resampled hourly, and we imputed missing values by first front filling within each episode and then filling any remaining gaps using the population mean. Continuous variables were z-standardised using the observed mean and standard deviation before imputation. A detailed summary of all EHR TS variables is available in the appendix.   

\begin{table}[h]
\caption{Overview of the two datasets. MIMIC‑IV is partitioned at the patient level; Physionet2012 uses predefined training/testing splits. Numbers in parentheses for ICU hours indicate the average episode length (hours) within each partition.}
\label{tab_dataset}
\resizebox{\columnwidth}{!}{%
\begin{tabular}{lcc}
\hline
Statistics & MIMIC-IV & Physionet2012 \\ \hline
Patients (training) & 23781 & 4000 \\
Patients (validation) & 5096 & N/A \\
Patients (testing) & 5096 & 4000 \\
ICU hours (training) & 4082k (145) & 192k (48) \\
ICU hours (validation) & 896k (146) & N/A \\
ICU hours (testing) & 883k (147) & 192k (48) \\
Mortality rate & 15.4\% & 14.0\% \\ \hline
\end{tabular}%
}
\end{table}

We extracted prescription orders, medication administrations, infusion orders, oxygen delivery settings, invasive line operations and renal replacement therapies (RRT/CRRT) settings as EHR events for an ICU episode. These events are extracted from corresponding tables and mapped to specific episodes via unique patient and ICU stay identifiers. The EHR events and EHR TS observations for the same ICU episode are synchronised using their recorded timestamps. We additionally processed some EHR events to LLM-friendly phrases by mapping all medications to ATC classifications and expanding ambiguous abbreviations into full English phrases.

\subsection{Experimental setting}
\label{result_3}
Similar to the ICU outcome prediction benchmark \citet{MIMICBenchmark}, we defined four downstream evaluation tasks based on ICU patient outcomes: (1) Mortality prediction uses the first 48 hours of measurements from an ICU episode to predict the risk of in-hospital mortality; (2) Phenotyping classifies episodes agains 25 acute phenotypes extracted from ICD-coded diagnoses; (3) Decompensation predicts immediate patient deteriorations within each episode, characterised by risk of organ dysfunctions; and (4) Remaining LoS is a time-to-event task to predict the days untile ICU discharge for a patient. Both decompensation and remaining LoS are rolling predictions, meaning predictions are made at regular intervals throughout each episode.

We compared LLM4EHR against supervised TS baselines and self‑supervised contrastive pre‑training methods for EHR TS modelling. For experiments on MIMIC-IV, we used the training set for self-supervised pre-training and supervised fine-tuning, the validation set for hyperparameter tuning and reported all models' performance on the test set. For additional experiments on Physionet2012, we used the challenge's pre-defined training set for k-shot adaptation and evaluated all models using the pre-defined testing set. All dataset splits were performed at the patient level, with Physionet2012 containing only the first $48$ hours of ICU stay per patient. Deep learning models were trained using the Adam optimiser \cite{kingma2014adam} on an Nvidia A100 GPU, and we reported performance across 10 random seeds for all models. We performed hyperparameter searches task-by-task for supervised models and used pre-training loss to determine the optimal hyperparameters for the self-supervised models. The hyperparameter search spaces for all baseline models are shown in \textbf{Table \ref{tab_baseline}}. 

\begin{table}[h]
\caption{Summary of the hyperparameter search space for the baseline models.}
\label{tab_baseline}
\resizebox{\columnwidth}{!}{%
\begin{tabular}{llll}
\hline
 & Hidden dim & Configuration & Specific parameters \\ \hline
LR & N/A & N/A & \begin{tabular}[c]{@{}l@{}}Inverse L1 strength:\\ {[}0.01, 0.1, 1.0, 3.0, 10.0{]}\end{tabular} \\ \hline
XGB & N/A & N/A & \begin{tabular}[c]{@{}l@{}}Max depth: \\ {[}3, 4, 5, 8, 10{]}\\ Subsample: \\ {[}0.2, 0.8, 1.0{]}\\ N estimators: \\ {[}300, 500{]}\end{tabular} \\ \hline
LSTM & 128, 256, 512 & \begin{tabular}[c]{@{}l@{}}N layers: \\ {[}1, 2, 3{]}\end{tabular} & N/A \\ \hline
GRU-D & 128, 256, 512 & N/A & N/A \\ \hline
T-TS/T-EHR & 256, 512 & \begin{tabular}[c]{@{}l@{}}N layers: \\ {[}4, 8{]}\\ N heads: \\ {[}4, 8{]}\end{tabular} & N/A \\ \hline
SimMTM & 256, 512 & \begin{tabular}[c]{@{}l@{}}N layers: \\ {[}4, 8{]}\\ N heads: \\ {[}4, 8{]}\end{tabular} & \begin{tabular}[c]{@{}l@{}}N neighbours: \\ {[}2, 3, 4{]}\\ Mask prob: \\ {[}0.1, 0.3, 0.5, 0.8{]}\\ Temperature: \\ {[}0.02, 0.05, 0.1{]}\end{tabular} \\ \hline
King et al. & 768 & \begin{tabular}[c]{@{}l@{}}N layers: \\ {[}4, 8{]}\\ N heads: \\ {[}4, 8{]}\end{tabular} & \begin{tabular}[c]{@{}l@{}}Mask prob: \\ {[}0.1, 0.3, 0.5{]}\\ Temperature: \\ {[}0.02, 0.05, 0.1{]}\end{tabular} \\ \hline
\end{tabular}%
}
\end{table}

For our implementation of LLM4EHR, we used a transformer with $8$ 
autoregressive transformer layers with $8$ attention heads per layer and a hidden dimension of $512$ as the EHR TS encoder. EHR summary embeddings were extracted from BioClinical ModernBERT \cite{BioClinicalModernBert}, which is a domain adapted version of ModernBERT \cite{ModernBERT} trained on medical texts. To allow more efficient LLM4EHR pre-training, the EHR embeddings were pre-computed and are projected into LLM4EHR's feature space via a dense projection layer, as described in \textbf{Section \ref{Sec::method_3}}. We implemented the baseline models as:

\textbf{(1) Linear Model (LR)}: Linear models included logistic regression for classifications and linear log‑normal accelerated failure time (AFT) model for remaining LoS predictions. Similar to \citet{lipton2015}, we extracted aggregated statistical features, including mean, minimum, maximum, standard deviation and count of measurements for each EHR TS variable. For sequence‑level classification, we extracted features over multiple clinically relevant windows: the 24 hours before the first heart rate measurement, the 24 and 48 hour windows following it, and the full interval up to the decision point. For rolling predictions, features were aggregated from the start of each patient trajectory up to the prediction point.

\textbf{(2) Gradient Boosting (XGB)}: We used the XGBoost implementation by \citet{xgboost}, with the same feature extraction method as the linear models. For classification, we trained the model using the binary logistic objective, and for LoS prediction, we used the log-normal AFT objective

\textbf{(3) LSTM}
The input to the LSTM consisted of 32 imputed and z-standardised EHR TS variables concatenated with 32 corresponding channel‑wise missingness indicator variables. We found that concatenating channel-wise missingness indicators with EHR TS variables yielded better single task performances than only using the EHR TS variables as input features. All LSTM models were trained end-to-end with task specific heads. For LoS prediction, we implemented the log-normal AFT objective in PyTorch \cite{Pytorch} following the implementation by \citet{weibullrnn}.

\textbf{(4) GRU-D}
We implemented the GRU-D by \citet{grud} as the second recurrent baseline. Unlike LSTM, GRU-D incorporates missingness indicators directly within the decayed GRU layer, allowing each partially observed variable to 'decay' towards its observed mean during training. GRU-D additionally uses time delta variables as input features, which are the time since the last observation for each EHR TS variable. Similar to LSTM, all GRU-D models were trained end-to-end, and we used the log-normal AFT objective for the remaining LoS prediction.

\textbf{(5) Supervised Transformers (T-TS \& T-EHR)}
We included two supervised transformer baselines, T‑TS and T‑EHR, as ablated versions of pre-trained LLM4EHR. Both models share the same autoregressive transformer structures and task heads with LLM4EHR, but are trained directly with task supervision. T-TS uses EHR TS variables as input, and T-EHR uses temporally aligned EHR summary embeddings as input. These models are conceptually similar to supervised transformer models for clinical TS, such as SAnD by \citet{SAnD}. Both models use the log-noraml AFT objective for LoS prediction.

\textbf{(6) SimMTM}
We implemented SimMTM by \citet{SimMTM} as a unimodal contrastive learning baseline. SimMTM introduces a contrastive reconstruction objective for masked TS modelling, where each masked time step is reconstructed via a weighted aggregation from $n$ neighbours. We implemented the SimMTM pre-training and aggregation objective with a bidirectional transformer encoder and pre-trained SimMTM on the training set. The input to SimMTM consists of 32 imputed and z-standardised EHR TS variables concatenated with 32 corresponding channel‑wise missingness indicator variables. The reconstruction loss was calculated only on the 32 EHR TS variables. Pre-trained SimMTM was adapted to downstream tasks using task specific heads and supervised fine-tuning on the training labels.

\textbf{(7) King et al.}
We implemented the pre-training method by \citet{king2023} as a multimodal contrastive learning baseline. The model by King et al. aligns EHR TS embeddings with clinical notes embeddings at the sequence level via a combination of cross-model contrastive \cite{InfoNCE} and masked reconstruction objectives. In our implementation, we used the same pre-training objective to align the embedding of the 32 EHR TS variables with clinical notes embeddings from BioClinical ModernBERT \cite{BioClinicalModernBert}. Since the model by King et al. did not include additional projection layers for clinical notes embeddings, we kept the hidden dimension of the pre-trained model the same as the LLM. Similar to SimMTM, the channel-wise missingness features were concatenated with the TS variables but were excluded from the reconstruction loss. The pre-trained models were then adapted to downstream tasks using task‑specific heads and supervised fine‑tuning on the training labels.

\subsection{Phenotyping}
Similar to the MIMIC-III \cite{MIMIC-III} benchmark by \citet{MIMICBenchmark}, the phenotyping task aims to predict the presence of 25 acute care conditions during an ICU stay. Labels were derived from ICD-coded diagnoses using the HCUP Clinical Classifications Software (CCS) mappings. For our experiments, we used the same 25 phenotypes as Harutyunyan et al., but extracted labels via the revised HCUP CCSR mappings for ICD-10 codes. We note that several broad phenotypes in CCS, such as 'Acute cerebrovascular disease', have been split into more specific subtypes in CSSR. In this case, our label sets cover a wider range of ICD-10 codes and therefore exhibit higher positive rates for some phenotypes. The prevalence of specific phenotypes is available in the appendix. Performance is summarised as macro-averaged AUROC and AUPRC across phenotype groups: All, Acute, Chronic/Mixed and low prevalence (Pr < 10\%). 

\begin{table*}[t]
\caption{Results for 25 phenotype predictions on MIMIC-IV with mean (std) over 10 runs,  best results are highlighted in bold, second best results are underlined, significant results ($p<0.05$ for paired t-test between best and second best) are noted with an asterisk.}
\label{tab_phenotype}
\resizebox{\textwidth}{!}{%
\begin{tabular}{c|cc|cc|cc|cc}
\hline
\multirow{2}{*}{Model} & \multicolumn{2}{c|}{All} & \multicolumn{2}{c|}{Acute} & \multicolumn{2}{c|}{Chronic/Mixed} & \multicolumn{2}{c}{Pr \textless 10\%} \\ \cline{2-9} 
 & AUROC & AUPRC & AUROC & AUPRC & AUROC & AUPRC & AUROC & AUPRC \\ \hline
LR & 0.776 (0.004) & 0.534 (0.007) & 0.782 (0.005) & 0.516 (0.011) & 0.768 (0.004) & 0.554 (0.006) & 0.702 (0.003) & 0.218 (0.005) \\
XGB & 0.805 (0.004) & 0.578 (0.005) & 0.809 (0.004) & 0.549 (0.005) & 0.801 (0.005) & 0.609 (0.006) & 0.734 (0.004) & 0.242 (0.006) \\
LSTM & 0.795 (0.011) & 0.542 (0.018) & 0.805 (0.009) & 0.520 (0.017) & 0.784 (0.011) & 0.565 (0.017) & 0.726 (0.012) & 0.209 (0.018) \\
GRU-D & 0.817 (0.004) & \underline{0.580 (0.006)} & \underline{0.826 (0.005)} & \underline{0.558 (0.007)} & 0.808 (0.005) & 0.603 (0.006) & \underline{0.750 (0.005)} & 0.248 (0.006) \\
T-TS & 0.796 (0.005) & 0.570 (0.006) & 0.802 (0.005) & 0.545 (0.007) & 0.790 (0.006) & 0.597 (0.006) & 0.729 (0.008) & 0.257 (0.009) \\
T-EHR & 0.803 (0.008) & 0.575 (0.011) & 0.809 (0.007) & 0.550 (0.009) & 0.797 (0.007) & 0.602 (0.011) & 0.737 (0.008) & 0.259 (0.010) \\
SimMTM & 0.811 (0.007) & 0.558 (0.010) & 0.815 (0.007) & 0.527 (0.008) & 0.806 (0.007) & 0.592 (0.0010) & 0.730 (0.008) & \underline{0.266 (0.012)} \\
King et al. & \underline{0.822 (0.005)} & \underline{0.580 (0.008)} & 0.825 (0.005) & 0.550 (0.006) & \underline{0.818 (0.006)} & \underline{0.612 (0.009)} & 0.749 (0.007) & 0.250 (0.008) \\
LLM4EHR & \textbf{0.844 (0.007)}$^{*}$ & \textbf{0.610 (0.009)}$^{*}$ & \textbf{0.848 (0.007)}$^{*}$ & \textbf{0.580 (0.008)}$^{*}$ & \textbf{0.840 (0.006)}$^{*}$ & \textbf{0.642 (0.010)}$^{*}$ & \textbf{0.771 (0.008)}$^{*}$ & \textbf{0.275 (0.009)}$^{*}$ \\ \hline
\end{tabular}%
}

\end{table*}

\begin{figure*}[h]
    \centering

    \begin{subfigure}[t]{0.48\textwidth}
        \centering
        \includegraphics[width=\linewidth]{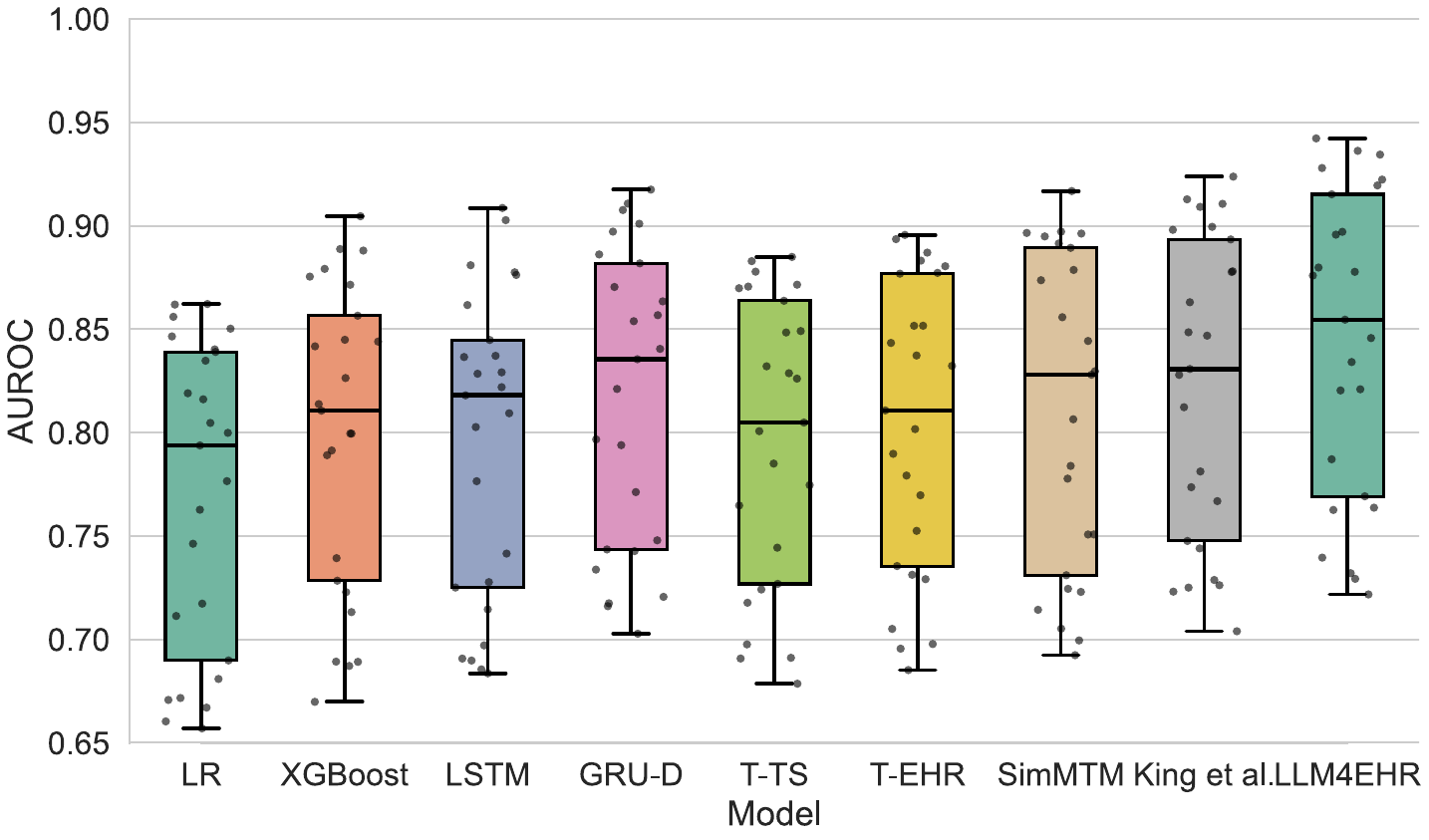}
        \caption{AUROC distribution by task}
        \label{fig:left_plot}
    \end{subfigure}
    \hfill
    \begin{subfigure}[t]{0.48\textwidth}
        \centering
        \includegraphics[width=\linewidth]{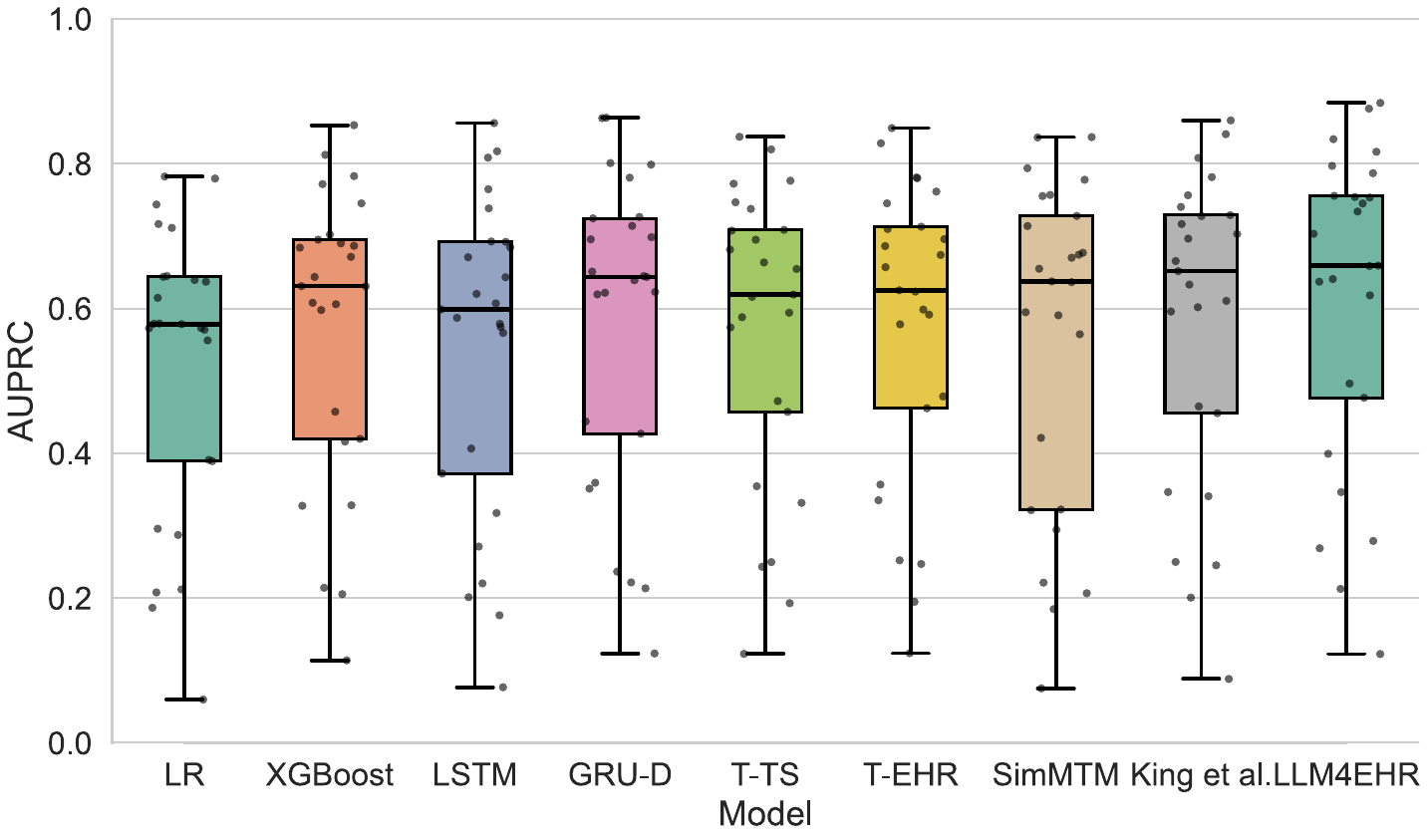}
        \caption{AUPRC distribution by task}
        \label{fig:right_plot}
    \end{subfigure}

    \caption{Box plot for mean AUROC and AUPRC across all 25 phenotypes.}
    \label{fig_pheno}
\end{figure*}

As seen in \textbf{Table \ref{tab_phenotype}}, LLM4EHR consistently outperforms all baselines in all phenotype groups. LLM4EHR achieved a $2.7\%$ ($+0.022$) gain in AUROC and a more sizeable $5.1\%$ ($+0.030$) gain in AUPRC over the next strongest baseline across the full set (All). The same pattern can be observed across the Acute and Chronic/Mixed phenotypes, indicating stable performance over clinically distinct phenotype subgroups. More importantly, LLM4EHR achieved significant gains in the low prevalence group (Pr < 10\%) with a $3.38\%$ improvement in AUPRC, suggesting that LLM4EHR is more robust to rare phenotypes.

However, as seen in \textbf{Figure \ref{fig_pheno}}, despite overall improvements, LLM4EHR's performance still varies wildly across specific phenotypes. This observation is expected, as some specific phenotypes like 'lower/upper respiratory', 'complications of surgical/medical care' or 'conduction disorders' are harder to predict from general EHR data or charted observations and often require specific data such as chest X-ray or ECG monitoring data to consistently predict. Finally, we note that the presence of Chronic/Mixed phenotypes was harder to predict than Acute phenotypes for all models, consistent with prior observations by Harutyunyan et al.

\subsection{Rolling predictions}

Rolling predictions include two longitudinal ICU outcome prediction tasks useful for patient management. We used the same two tasks as \citet{MIMICBenchmark} but defined the labels differently. Rolling decompensation predictions aim to predict imminent physiological deterioration. While Harutyunyan et al. used in-unit mortality as a proxy label, we instead use changes in SOFA score \cite{SOFA, SOFAdev}, which more directly reflect acute organ dysfunction for patients under intensive care. Specifically, we computed SOFA score hourly using the worst measurements over the past 24 hour window. Missing components are treated as 0 (no related risk) if no previous observations and front filled (no change in risk) if previous observations exist. At time $t$, decompensation is marked by either patient death or a change over $+2$ in SOFA score. The change is SOFA score is calculate as the difference between SOFA at $t$ and SOFA at $t-1$. At each prediction point $t_{p}$, we locate the nearest decompensation indicator at $t'>t_{p}$ within the same ICU stay and assign a positive label to $t_{p}$ if $t' - t_{p} <= 24$. We report AUROC, AUPRC and F1 score for all models.

\begin{table}[]
\caption{Summary of Decompensation and LoS labels across downsampled prediction points}
\label{tab_rolling_stat}
\resizebox{0.8\columnwidth}{!}{%
\begin{tabular}{cccc}
\hline
 & Training & Validation & Testing \\ \hline
\begin{tabular}[c]{@{}c@{}}Prediction\\ Points\end{tabular} & 525k & 117k & 115k \\
\begin{tabular}[c]{@{}c@{}}Decomp\\ Rate\end{tabular} & 20.67\% & 20.93\% & 20.2\% \\
\begin{tabular}[c]{@{}c@{}}LoS censor\\ Rate\end{tabular} & 3.97\% & 4.74\% & 4.60\% \\
LoS Mean & 8.48 & 8.90 & 10.91 \\
LoS Max & 158.43 & 110.80 & 369.75 \\
LoS Median & 4.95 & 5.22 & 5.09 \\
LoS IQR & 7.94 & 8.54 & 8.44 \\
LoS Skew & 4.19 & 3.18 & 8.86 \\ \hline
\end{tabular}%
}
\end{table}

For the remaining LoS prediction, we avoided the task definition by Harutyunyan et al. and formulated the task as time-to-event regression with an accelerated failure time objective. More specifically, we assume that covariates (EHR TS variables) would hasten or prolong the patient's discharge from ICU. The aim is to predict days until discharge at at each prediction point and we imposed right censoring at 30 days. Given that the utility of LoS prediction is mostly for ICU bed management, prediction over arbitrarily long horizon is not clinically useful due the presence of extreme outliers (\textbf{Table \ref{tab_rolling_stat}}). In our case, we consider 30 days to be a generous horizon as less than $5\%$ of the prediction points are censored. Since the censor rate is low, we report MAE and R2 for uncensored predictions and C-index for all predictions. Due to the extreme number of prediction points (\textbf{Table \ref{tab_dataset}}), we downsampled the rolling prediction to be one prediction every 5 hours. We additionally limited the prediction period to be from 24 hours after ICU admission and 24 hours before ICU discharge to avoid trivial decompensation and LoS predictions.

\begin{table*}[h]
\caption{Results for rolling decompensation and LoS predictions on MIMIC-IV with mean (std) over 10 runs, best results are highlighted in bold, second best results are underlined, significant results ($p<0.05$ for paired t-test between best and second best) are noted with an asterisk.}
\label{tab_rolling}
\resizebox{0.8\textwidth}{!}{%
\begin{tabular}{c|ccc|ccc}
\hline
\multirow{2}{*}{Models} & \multicolumn{3}{c|}{Decompensation} & \multicolumn{3}{c}{LoS} \\ \cline{2-7} 
 & AUROC & AUPRC & F1 & MAE & R2 & C-index \\ \hline
LR & 0.635 (0.005) & 0.296 (0.007) & 0.370 (0.011) & 2.630 (0.006) & 0.058 (0.008) & 0.637 (0.005) \\
XGB & 0.701 (0.006) & 0.407 (0.007) & 0.194 (0.009) & 4.052 (0.019) & 0.093 (0.011) & 0.652 (0.007) \\
LSTM & 0.777 (0.008) & 0.561 (0.011) & 0.515 (0.018) & 2.574 (0.040) & 0.036 (0.027) & 0.608 (0.008) \\
GRU-D & 0.848 (0.007) & 0.669 (0.009) & 0.586 (0.012) & \underline{2.352 (0.030)} & \underline{0.216 (0.021)} & 0.653 (0.010) \\
T-TS & \underline{0.864 (0.008)} & \underline{0.691 (0.009)} & \underline{0.623 (0.015)} & 2.356 (0.029) & 0.148 (0.024) & 0.658 (0.009) \\
T-EHR & 0.808 (0.008) & 0.587 (0.011) & 0.500 (0.014) & 2.486 (0.033) & 0.132 (0.026) & 0.646 (0.009) \\
SimMTM & 0.845 (0.008) & 0.658 (0.009) & 0.572 (0.015) & 2.380 (0.019) & 0.200 (0.024) & \underline{0.659 (0.011)} \\
King et al. & 0.768 (0.007) & 0.516 (0.008) & 0.378 (0.009) & 2.651 (0.027) & 0.145 (0.020) & 0.656 (0.008) \\
LLM4EHR & \textbf{0.876 (0.007)}$^{*}$ & \textbf{0.723 (0.009)}$^{*}$ & \textbf{0.632 (0.012)} & \textbf{2.343 (0.026)} & \textbf{0.226 (0.024)} & \textbf{0.675 (0.009)} \\ \hline
\end{tabular}%
}
\end{table*}

As shown in \textbf{Table \ref{tab_rolling}}, LLM4EHR achieves the best performance across both rolling prediction tasks, with significant gains in rolling decompensation. Since our definition of decompensation assumes that physiological decline can occur at any point within the prediction window, the task requires models to process longitudinal observations to maintain stable performance across all metrics, as seen in models such as LR and XGB achieving substantially lower F1 scores compared with deep learning baselines. In this case, LLM4EHR's consistent performance gain over the next best model in all metrics suggests that pre-training helped LLM4EHR to better capture temporal dependencies during downstream tuning, which is further supported by LLM4EHR outperforming the ablated baselines T-TS and T-EHR.

In contrast, the performance differences between other baseline models and LLM4EHR in rolling LoS prediction are less significant. LLM4EHR performed on par with GRU-D and SimMTM in all metrics, with only marginal performance gains. The overall R2 was poor across all models, consistent with prior work on ICU LoS regression, with \citet{losmimic} achieving an R2 of $0.24$ on a carefully curated MIMIC-IV cohort with outliers removed. This outcome demonstrates that it is far easier to discriminate between short and prolonged stays than to predict the exact remaining time to discharge. However, LLM4EHR noticeably outperformed the ablated baselines T-TS and T-EHR, suggesting that LLM4EHR benefited from semantic alignment pre-training even in a complex downstream task with high label variability.

\subsection{Mortality prediction}
\label{sec::mortality}
\begin{table}[h]
\caption{Results for full-shot mortality prediction on MIMIC-IV and Physionet2012 with mean (std) over 10 runs, best results are highlighted in bold, second best results are underlined, significant results ($p<0.05$ for paired t-test between best and second best) are noted with an asterisk.}
\label{tab_mortality}
\resizebox{\columnwidth}{!}{%
\begin{tabular}{c|cc|cc}
\hline
\multirow{2}{*}{Model} & \multicolumn{2}{c|}{MIMIC-IV} & \multicolumn{2}{c}{Physionet2012} \\ \cline{2-5} 
 & AUROC & AUPRC & AUROC & AUPRC \\ \hline
LR & 0.824 (0.004) & 0.354 (0.003) & 0.754 (0.003) & 0.322 (0.006) \\
XGB & 0.841 (0.003) & 0.403 (0.005) & 0.827 (0.004) & 0.398 (0.005) \\
LSTM & 0.835 (0.007) & 0.487 (0.006) & 0.804 (0.007) & 0.443 (0.008) \\
GRU-D & 0.850 (0.009) & 0.533 (0.012) & \underline{0.843 (0.008)} & 0.484 (0.010) \\
T-TS & 0.844 (0.008) & 0.511 (0.009) & 0.839 (0.008) & 0.452 (0.010) \\
T-EHR & 0.839 (0.011) & 0.509 (0.012) & N/A & N/A \\
SimMTM & 0.852 (0.007) & 0.522 (0.009) & 0.841 (0.008) & \underline{0.487 (0.009)} \\
King et al. & \underline{0.857 (0.007)} & \underline{0.534 (0.008)} & 0.829 (0.007) & 0.484 (0.009) \\
LLM4EHR & \textbf{0.870 (0.008)}$^{*}$ & \textbf{0.542 (0.009)} & \textbf{0.856 (0.006)}$^{*}$ & \textbf{0.495 (0.008)}$^{*}$ \\ \hline
\end{tabular}%
}
\end{table}

\begin{figure}[]
  \centering
  \includegraphics[width=\columnwidth]{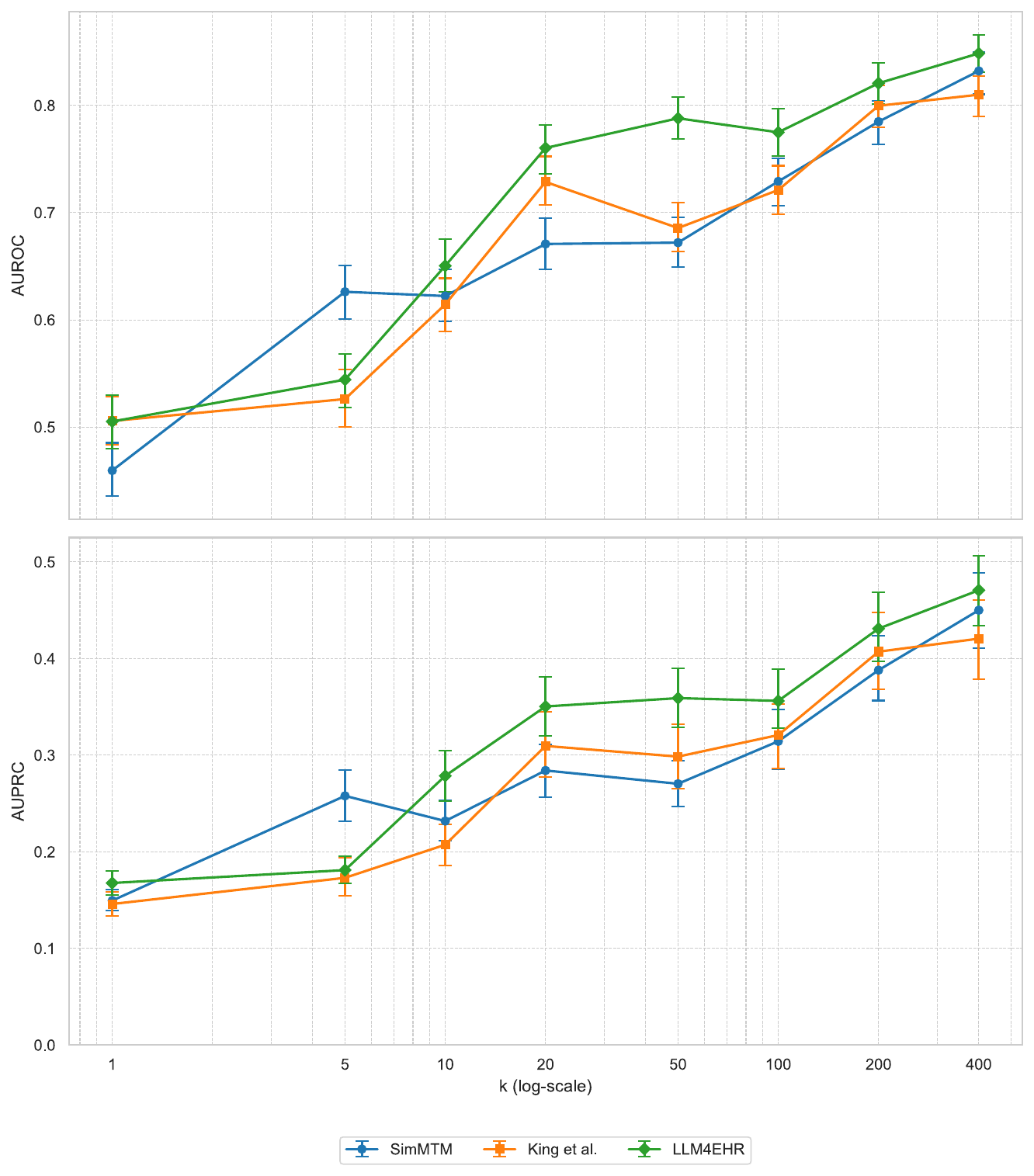}
  \caption{AUROC (upper) and AUPRC (lower) for k-shot mortality prediction on Physionet2012 with mean and $95\%$ confidence over 1000 bootstrap samples.}
  \label{fig_k_shot}
\end{figure}

Mortality prediction aims to predict in-hospital mortality using the first 48 hours of ICU observations. We used the Physionet2012 challenge's definition of in-hospital mortality and assigned positive labels to patients who died before hospital discharge after spending at least 48 hours in the ICU \cite{Physionet2012}. We report AUROC and AUPRC for all models in a full‑shot setting on both MIMIC‑IV and Physionet2012. Additionally, we used k-shot adaptation to assess the generalisability and label efficiency of pre-trained models. In this case, models pre-trained using MIMIC-IV episodes (SimMTM, the model by King et al. and LLM4EHR) were fine-tuned using $k$ positive and $k$ negative samples from the Physionet2012 training set and tested on the full test set. For each $k$, we report the mean AUROC and AUPRC with $95\%$ confidence intervals using 1000 bootstrap samples.

The full-shot mortality prediction results in \textbf{Table \ref{tab_mortality}} show that models pre-trained on MIMIC-IV patient TS outperformed supervised models on MIMIC-IV mortality prediction, with LLM4EHR achieving the best overall performance in both datasets. LLM4EHR experienced a larger relative AUROC drop ($-1.6\%$) than GRU-D and SimMTM when adapted to Physionet2012, likely caused by LLM4EHR capturing MIMIC-IV specific patterns during pre-training. This observation is additionally supported by the few-shot results in \textbf{Figure \ref{fig_k_shot}}, where LLM4EHR performed worse than SimMTM in extremely low-k settings $k<5$ but overtook both SimMTM and the model by King et al. at $k=10$ and maintained a steady lead during moderate and high-k settings. The model by King et al., similarly pre-trained via cross-modal alignment on MIMIC-IV, experienced a larger relative performance drop than LLM4EHR ($-3.4\%$ in AUROC) when adapted to Physionet2012, indicating that our temporal alignment objective is less sensitive to data shift than the stationary alignment objective used by King et al.

\subsection{Further analysis of LLM4EHR}
\label{sec::ablation}

\subsubsection*{Ablation study of the pre-training objective}
We experimented with different pre-training objectives to assess the impact of each component within LLM4EHR's semantic alignment objective. As shown in \textbf{Table \ref{tab_ablation_1}}, the largest performance gain comes from the combination of $\mathcal{L}_{recon}$ with $\mathcal{L}_{NCE}$, with the addition of $\mathcal{L}_{\omega}$ delivering marginal but consistent performance gains across all four downstream tasks. The results demonstrate that temporal alignment is crucial for LLM4EHR, with $\mathcal{L}_{\omega}$ having a regularising effect on the learned TS embeddings. Additionally, models pre-trained without the reconstruction objective are significantly worse at all tasks, suggesting that reconstruction is a necessary auxiliary task to contrastive alignment.

\begin{table*}[]
\caption{Ablation study on MIMIC-IV downstream tasks using different pre-training objectives, results are reported as mean (std) over 10 runs, best results are highlighted in bold, second best results are underlined, significant results ($p<0.05$ for paired t-test between best and second best) are noted with an asterisk.}
\label{tab_ablation_1}
\resizebox{\textwidth}{!}{%
\begin{tabular}{l|cc|cc|cc|cc}
\hline
\multirow{2}{*}{Objective} & \multicolumn{2}{c|}{Mortality} & \multicolumn{2}{c|}{Phenotyping} & \multicolumn{2}{c|}{Decompensation} & \multicolumn{2}{c}{LoS} \\ \cline{2-9} 
 & AUROC & AUPRC & \begin{tabular}[c]{@{}c@{}}AUROC\\ (marco)\end{tabular} & \begin{tabular}[c]{@{}c@{}}AUPRC\\ (marco)\end{tabular} & AUROC & AUPRC & MAE & R2 \\ \hline
$\mathcal{L}_{recon}$ & 0.847 (0.006) & 0.520 (0.009) & 0.809 (0.008) & 0.554 (0.009) & 0.832 (0.009) & 0.662 (0.013) & 2.382 (0.021) & 0.209 (0.019) \\
$\mathcal{L}_{NCE}$ & 0.836 (0.010) & 0.499 (0.013) & 0.786 (0.009) & 0.540 (0.013) & 0.806 (0.008) & 0.580 (0.011) & 2.582 (0.031) & 0.099 (0.033) \\
$\mathcal{L}_{recon}$ + $\mathcal{L}_{NCE}$ & \underline{0.863 (0.008)} & \underline{0.540 (0.010)} & \underline{0.839 (0.007)} & \underline{0.607 (0.009)} & \underline{0.868 (0.008)} & \underline{0.712 (0.009)} & \textbf{2.340 (0.022)} & \underline{0.222 (0.023)} \\
$\mathcal{L}_{align}$ & \textbf{0.870 (0.008)}$^{*}$ & \textbf{0.542 (0.009)} & \textbf{0.844 (0.007)}$^{*}$ & \textbf{0.610 (0.009)} & \textbf{0.876 (0.007)}$^{*}$ & \textbf{0.723 (0.009)}$^{*}$ & \underline{2.343 (0.026)} & \textbf{0.226 (0.024)} \\ \hline
\end{tabular}%
}
\end{table*}

\begin{table*}[]
\caption{Ablation study on MIMIC-IV downstream tasks using different LLMs for embedding EHR summaries, results are reported as mean (std) over 10 runs, best results are highlighted in bold, second best results are underlined, significant results ($p < 0.05$ for paired t-test between best and second best) are noted with an asterisk.}
\label{tab_ablation_2}
\resizebox{\textwidth}{!}{%
\begin{tabular}{l|cc|cc|cc|cc}
\hline
\multirow{2}{*}{} & \multicolumn{2}{c|}{Mortality} & \multicolumn{2}{c|}{Phenotype} & \multicolumn{2}{c|}{Decompensation} & \multicolumn{2}{c}{LoS} \\ \cline{2-9} 
 & AUROC & AUPRC & \begin{tabular}[c]{@{}c@{}}AUROC\\ (macro)\end{tabular} & \begin{tabular}[c]{@{}c@{}}AUPRC\\ (macro)\end{tabular} & AUROC & AUPRC & MAE & R2 \\ \hline
ModernBERT & 0.850 (0.009) & 0.530 (0.009) & 0.822 (0.006) & 0.577 (0.009) & 0.869 (0.010) & \underline{0.715 (0.009)} & 2.348 (0.031) & \underline{0.215 (0.026)} \\
BioClinical ModernBERT & \textbf{0.870 (0.008)} & \textbf{0.542 (0.009)} & \textbf{0.844 (0.007)}$^{*}$ & \textbf{0.610 (0.009)}$^{*}$ & \textbf{0.876 (0.007)} & \textbf{0.723 (0.009)}$^{*}$ & \textbf{2.343 (0.026)} & \textbf{0.226 (0.024)} \\
Longformer & 0.848 (0.008) & 0.529 (0.012) & 0.827 (0.010) & 0.582 (0.015) & 0.866 (0.008) & 0.708 (0.013) & \underline{2.344 (0.022)} & 0.208 (0.018) \\
Clinical-Longformer & \underline{0.864 (0.007)} & \underline{0.541 (0.011)} & \underline{0.839 (0.011)} & \underline{0.583 (0.017)} & \underline{0.869 (0.012)} & 0.707 (0.015) & 2.344 (0.023) & 0.210 (0.018) \\
RoBERTa & 0.838 (0.011) & 0.392 (0.013) & 0.810 (0.013) & 0.552 (0.021) & 0.823 (0.006) & 0.660 (0.008) & 2.653 (0.020) & 0.060 (0.021) \\
BioMed-RoBERTa & 0.842 (0.009) & 0.395 (0.014) & 0.818 (0.009) & 0.572 (0.011) & 0.829 (0.009) & 0.658 (0.012) & 2.650 (0.021) & 0.072 (0.019) \\ \hline
\end{tabular}%
}
\end{table*}

\begin{table}[]
\caption{Summary of LLMs for embedding EHR summaries}
\label{tab_llm}
\resizebox{0.8\columnwidth}{!}{%
\begin{tabular}{lcc}
\hline
 & \begin{tabular}[c]{@{}c@{}}Context\\ Length\end{tabular} & Size \\ \hline
ModernBERT & 8192 & 150M \\
BioClinical ModernBERT & 8192 & 150M \\
Longformer & 4096 & 150M \\
Clinical-Longformer & 4096 & 150M \\
RoBERTa & 512 & 125M \\
BioMed-RoBERTa & 512 & 125M \\ \hline
\end{tabular}%
}
\end{table}

\begin{table*}[]
\caption{The effects of the contrastive temperature $\tau$ on the downstream task, since the optimal $\tau$ was different for each task, we selected $\tau = 0.02$ for all evaluations as it achieved competitive results in all three tasks.}
\label{tab_ablation_3}
\resizebox{\textwidth}{!}{%
\begin{tabular}{c|cc|cc|cc|cc}
\hline
\multirow{2}{*}{} & \multicolumn{2}{c|}{Mortality} & \multicolumn{2}{c|}{Phenotype} & \multicolumn{2}{c|}{Decompensation} & \multicolumn{2}{c}{LoS} \\ \cline{2-9} 
 & AUROC & AUPRC & \begin{tabular}[c]{@{}c@{}}AUROC\\ (macro)\end{tabular} & \begin{tabular}[c]{@{}c@{}}AUPRC\\ (macro)\end{tabular} & AUROC & AUPRC & MAE & R2 \\ \hline
$\tau = 0.02$ & \textbf{0.870 (0.008)}$^{*}$ & \textbf{0.542 (0.009)} & \underline{0.844 (0.007)} & \textbf{0.610 (0.009)} & \textbf{0.876 (0.007)}$^{*}$ & \textbf{0.723 (0.009)}$^{*}$ & \underline{2.343 (0.026)} & \underline{0.226 (0.024)} \\
$\tau = 0.08$ & \underline{0.862 (0.008)} & \underline{0.537 (0.010)} & \textbf{0.846 (0.008)} & \underline{0.609 (0.011)} & \underline{0.856 (0.007)} & \underline{0.699 (0.010)} & 2.351 (0.022) & \textbf{0.228 (0.020)} \\
$\tau = 0.14$ & 0.849 (0.007) & 0.529 (0.009) & 0.811 (0.007) & 0.592 (0.012) & 0.825 (0.005) & 0.657 (0.007) & \textbf{2.339 (0.017)} & 0.208 (0.021) \\
$\tau = 0.20$ & 0.852 (0.008) & 0.531 (0.011) & 0.789 (0.006) & 0.571 (0.008) & 0.805 (0.006) & 0.621 (0.009) & 2.427 (0.025) & 0.152 (0.022) \\
$\tau = 0.50$ & 0.844 (0.012) & 0.522 (0.016) & 0.779 (0.007) & 0.554 (0.010) & 0.767 (0.010) & 0.576 (0.017) & 2.554 (0.031) & 0.112 (0.029) \\ \hline
\end{tabular}%
}
\end{table*}

\subsubsection*{The choice of LLMs}
Since our pre-training framework is compatible with most LLM structures, we investigated the impacts of using different LLMs to embed EHR summaries. We compared the results obtained from BioClinical ModernBERT \cite{BioClinicalModernBert} against two other LLMs trained on medical texts: Clinical-Longformer \cite{ClinicalLongformer} and BioMed-RoBERTa \cite{clinicalRoberta}. Additionally, we also include results for ModernBERT \cite{ModernBERT}, Longformer \cite{Longformer} and RoBERTa \cite{RoBERTa}. \textbf{Table \ref{tab_llm}} summarises the context length and size of the LLMs. \textbf{Table \ref{tab_ablation_2}} shows that LLM4EHR pre-trained with long context LLMs outperformed LLM4EHR pre-trained with short context LLMs. BioClinical ModernBERT and Clinical-Longformer consistently yielded better performance in downstream evaluations than their general-purpose counterparts, indicating that domain-adapted LLMs are crucial for LLM4EHR pre-training. This observation, along with the ablation results in \textbf{Table \ref{tab_ablation_1}}, shows that our pre-training objective facilitated meaning transfer of semantic knowledge from pre-trained LLM to LLM4EHR embedded EHR TS features.

\subsubsection*{The effects of the temperature $\boldsymbol{\tau}$}
The temperature $\tau$ is the scaling term for the similarity matrices in both $\mathcal{L}_{NCE}$ and $\mathcal{L}_{\omega}$, and it is an important hyperparameter for LLM4EHR pre-training. Intuitively, a small $\tau$ sharpens the differences between similar and dissimilar samples, forcing the model to distinguish positive pairs from negatives aggressively. As we formulated $\mathcal{L}_{\omega}$ to be a regularisation term for the main alignment loss $\mathcal{L}_{NCE}$, we prioritised investigating the effects of $\tau$ in $\mathcal{L}_{NCE}$ and kept $\tau = 0.14$ for $\mathcal{L}_{\omega}$. In practice, we found that using $\tau<0.08$ for $\mathcal{L}_{\omega}$ causes $\mathcal{L}_{align}$ to be dominated by $\mathcal{L}_{\omega}$, leading to unstable pre-training. \textbf{Table \ref{tab_ablation_3}} shows that a smaller $\tau$ for $\mathcal{L}_{NCE}$ is overall better for pre-training, with a noticeable drop in performance between $\tau = 0.08$ and $\tau = 0.14$.

\section{Discussion}
\label{Sec_5}
We presented LLM4EHR as a multimodal pre-training framework for clinical foundation models. The results in \textbf{Section \ref{Sec_4}} show that by jointly modelling EHR TS with EHR events, LLM4EHR achieves state-of-the-art performance on clinically significant downstream tasks, including mortality prediction, patient phenotyping, ICU decompensation, and remaining LoS prediction. The cross-dataset mortality prediction results in \textbf{Section \ref{sec::mortality}} show that pre-trained LLM4EHR can efficiently generalise to new ICU patient cohorts via few-shot learning. The ablation studies in \textbf{Section \ref{sec::ablation}} show that our $\omega$-regularised alignment objective improves the consistency of the learned EHR TS embedding and that using domain adapted LLM for embedding EHR event summaries during pre-training was crucial for LLM4EHR. 

The major limitation of LLM4EHR is scalability to larger language models, which have considerable memory requirements to train. Future research could consider more efficient contrastive learning structures such as \citet{he2020momentum}, which uses a memory queue to store past samples for calculating the contrastive objective. Additionally, we focused on intensive care episodes during pre-training; future studies should consider modelling extended patient trajectories spanning multiple stays, which would extend LLM4EHR to additional downstream tasks useful for long-term care monitoring.  Lastly, we evaluated the performance of LLM4EHR using learned TS embeddings, given our focus on predicting ICU patient outcomes. Future research should investigate the applicability of LLM4EHR with schema alignment-based pre-training \cite{Sun2024} for natural language inference, such as question answering with EHR patient observations.

\printcredits

\section*{Conflict of interest}
The authors declare that they have no known competing financial interests or personal relationships that could have appeared to influence the work reported in this paper.

\section*{Acknowledgments}
This work was supported by the Great Ormond Street Hospital (GOSH) Charity (grant number 21PP30). PB would like to acknowledge funding via the Royal Academy of Engineering and Great Ormond Street Hospital, the UK Dementia Research Institute (award number UK DRI-7002) through UK DRI Ltd, principally funded by the Medical Research Council.

\section*{Data availability}
The MIMIC-IV and Physionet Challenge 2012 are publicly available. All code used for curating the dataset and running our experiments can be found at: 
\url{https://github.com/CrankyWilliam/LLM4EHR}.

\bibliographystyle{model1-num-names}

\bibliography{main}

\newpage
\appendix
\section{Appendix}
\label{sec:appendix}

\subsection{Curating the datasets}
Here we describe the process for curating the MIMIC-IV dataset, corresponding to the Python implementation provided in our code repository. Since we did not further filter or process the Physionet Challenge 2012 dataset, we refer to the original implementation by \citet{Physionet2012} for details regarding dataset curation and cohort selection. 

\subsubsection{MIMIC-IV cohort}
\begin{figure*}[h]
\centering
\includegraphics[width=.8\linewidth]{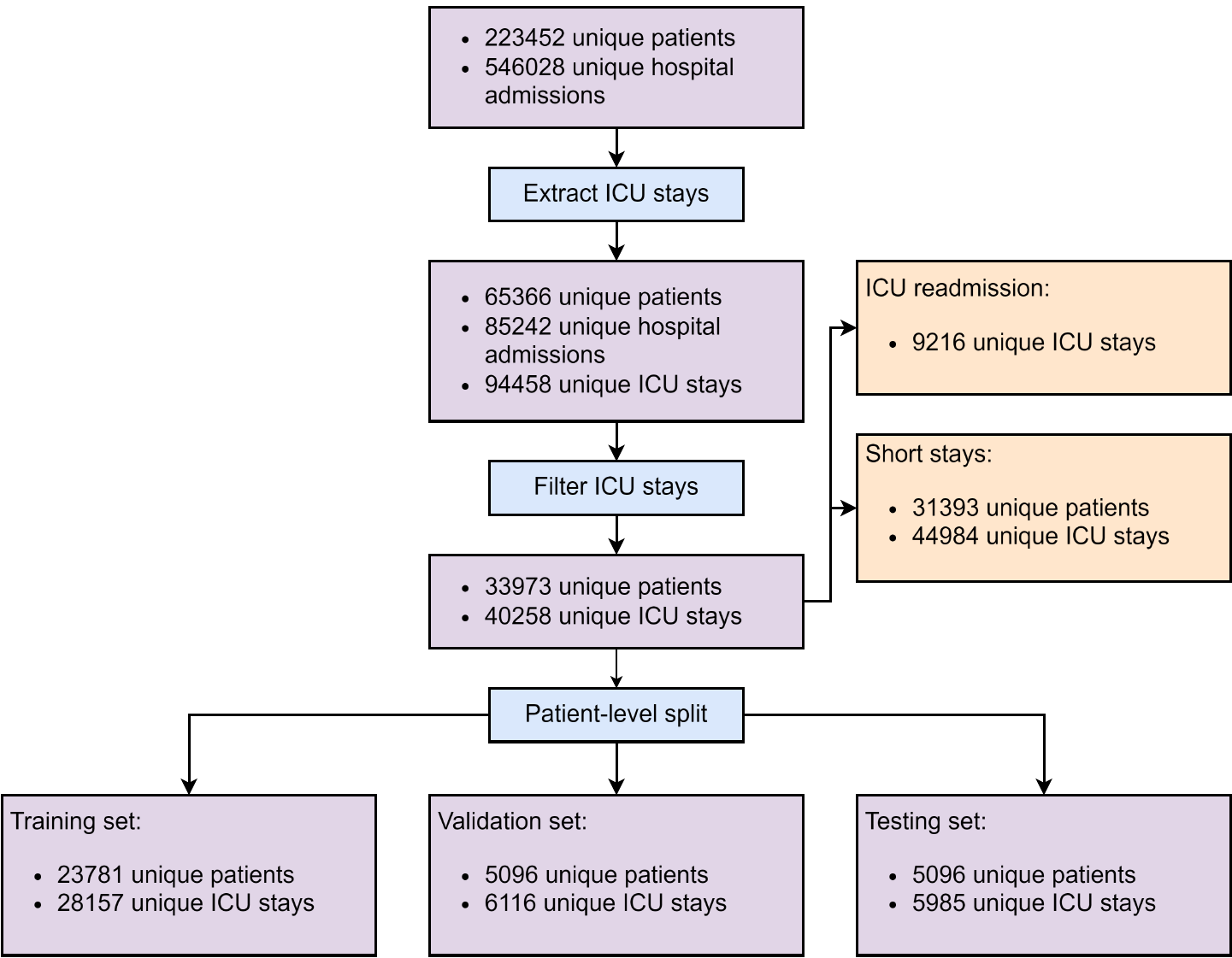}
\caption{Flowchart for curating the MIMIC-IV ICU stay patient cohort, short stays $<48$ hours are removed to allow adequate patient observations before prediction.}
\label{fig_cohort}
\end{figure*}

We curated a cohort of patients with at least one recorded ICU stay. Different from the cohort selection process by \citet{MIMICBenchmark}, we included patients with multiple ICU stays per hospital admission but kept only the first ICU stay per hospital admission. \citet{MIMICBenchmark} argues that including hospital admissions with multiple ICU stays makes the mortality and phenotyping labels ambiguous. While doing so would be valid for predicting the onset of adverse events in a unit, such as predicting the onset of cardiac arrest in the ICU, we found no reason to remove hospital admissions with multiple ICU stays, as both phenotype and mortality labels are defined for unique hospital admissions rather than specific stays. The labels for phenotype and mortality would still be valid for a given ICU stay if a patient was discharged then subsequently readmitted to ICU provided both stays occurs within the same hospital admission period. Therefore, we elect to retain the first ICU stay for each hospital admission. For direct transfers between ICU wards, when the discharge time of the first ICU ward stay overlaps with the admission time of the second ICU ward stay, we merge the two stays into a single long stay. The process of curating MIMIC-IV cohort is shown in \textbf{Figure \ref{fig_cohort}}.

\subsubsection{MIMIC-IV Mortality labels}

We used the Physionet2012 Challenge's definition of in-hospital mortality and assigned positive labels to patients who died before hospital discharge after spending at least 48 hours in the ICU. Since we only included stays longer than 48 hours, all curated MIMIC-IV episodes are eligible for mortality prediction. Here, we specify that the patient mortality label is associated with unique hospital stays rather than unique ICU stays, as the hospital discharge time was used to determine in-hospital patient mortality. In this case, we included only the first ICU stay for each hospital admission to avoid duplicating mortality labels.

\subsubsection{MIMIC-IV Phenotype labels}

Similar to the MIMIC-III benchmark by \citet{MIMICBenchmark},c the phenotyping task aims to predict the presence of 25 acute care conditions during an ICU stay. Labels were derived from ICD-coded diagnoses using the HCUP Clinical Classifications Software (CCS) mappings. In their original implementation Harutyunyan et al. noted that phenotype labels are associated with hospital stays rather than ICU stays and are predicted retrospectively using an entire ICU episode. For our experiments, we used the same 25 phenotypes as Harutyunyan et al., but extracted labels via the revised HCUP CCSR mappings for ICD-10 codes. We note that several broad phenotypes in CCS, such as 'Acute cerebrovascular disease', have been split into more specific subtypes in CSSR. \textbf{Table \ref{tab_phenotype_label}} summarises the 25 phenotype labels in MIMIC-IV.

\begin{table*}[t]
\caption{Summary of 25 phenotype labels in MIMIC-IV.}
\label{tab_phenotype_label}
\resizebox{0.6\textwidth}{!}{%
\begin{tabular}{ccc}
\hline
Phenotype & Type & Prevalence \\ \hline
Acute cerebrovascular disease & Acute & 0.159 \\
Acute myocardial infarction & Acute & 0.110 \\
Acute renal failure & Acute & 0.375 \\
Cardiac dysrhythmias & Chronic/Mixed & 0.375 \\
Chronic obstructive pulmonary disease & Chronic/Mixed & 0.175 \\
Complications of surgical/medical care & Acute & 0.099 \\
Conduction disorders & Chronic/Mixed & 0.097 \\
Congestive heart failure; nonhypertensive & Chronic/Mixed & 0.312 \\
Coronary atherosclerosis and related & Chronic/Mixed & 0.334 \\
Diabetes mellitus with complications & Chronic/Mixed & 0.151 \\
Diabetes mellitus without complications & Chronic/Mixed & 0.173 \\
Disorder of lipid metabolism & Chronic/Mixed & 0.433 \\
Essential hypertension & Chronic/Mixed & 0.404 \\
Fluid and electrolyte disorder & Acute & 0.498 \\
Gastrointestinal hemorrhage & Acute & 0.092 \\
Hypertension with complications & Chronic/Mixed & 0.267 \\
Other liver disease & Chronic/Mixed & 0.122 \\
Other lower respiratory disease & Acute & 0.093 \\
Other upper respiratory disease & Acute & 0.028 \\
Pleurisy; pneumothorax; pulmonary collapse & Acute & 0.169 \\
Pheumonia & Acute & 0.245 \\
Respiratory failure; insufficiency; arrest & Acute & 0.386 \\
Septicemia (except in labour) & Acute & 0.219 \\
Shock & Acute & 0.261 \\ \hline
\end{tabular}%
}
\end{table*}

\subsubsection{MIMIC-IV decompensation labels}

Rolling decompensation predictions aim to predict imminent physiological deterioration. While \citet{MIMICBenchmark} used in-unit mortality as a proxy label, we instead use changes in SOFA score, which more directly reflect acute organ dysfunction. Specifically, we computed SOFA score hourly using the worst measurements over the past 24 hour window. Missing components are treated as 0 (no related risk) if no previous observations and front filled (no change in risk) if previous observations exist. At time $t$, decompensation is marked by either patient death or a change over $+2$ in SOFA score. The change is SOFA score is calculate as the difference between SOFA at $t$ and SOFA at $t-1$. At each prediction point $t_{p}$, we locate the nearest decompensation indicator at $t'>t_{p}$ within the same ICU stay and assign a positive label to $t_{p}$ if $t' - t_{p} <= 24$. SOFA score is calculated using the criteria in \textbf{Table \ref{tab_sofa_score}}.

\begin{table*}[]
\caption{Scoring criteria for SOFA score components summarised from \citet{SOFA}, with additional Norepinephrine equivalent dosages from \citet{SOFAdev}}
\label{tab_sofa_score}
\resizebox{\textwidth}{!}{%
\begin{tabular}{ccccccc}
\hline
 & Respiratory system & Nervous system & Cardiovascular System & Liver system & Coagulation & Kidneys \\ \hline
SOFA score & PaO2/FiO2 (mmHg) & Glasgow coma score & MAP/Vasopressor & Bilirubin (mg/dl) {[}umo/L{]} & Platelets x1000/ml & \begin{tabular}[c]{@{}c@{}}Creatinine (mg/dl); \\ Urine output (ml/day)\end{tabular} \\
0 & \textgreater 400 & 15 & MAP \textgreater 70 mmHg & \textless 1.2 (\textless 20) & \textgreater 150 & \textless 1.2 \\
1 & \textless 400 & 13-14 & MAP \textless 70 mmHg & 1.2 - 1.9 {[}20 - 32{]} & \textless 150 & 1.2 - 1.9 \\
2 & \textless 300 & 10-12 & \begin{tabular}[c]{@{}c@{}}Dopamine \textless{}= 5 ug/kg/min\\ Dobutamine (any dosage)\end{tabular} & 2.0 - 5.9 {[}33 - 101{]} & \textless 100 & 2.0 - 3.4 \\
3 & \begin{tabular}[c]{@{}c@{}}\textless 200 with respiratory \\ support\end{tabular} & 6-9 & \begin{tabular}[c]{@{}c@{}}Norepinephrine \textless{}= 0.1 ug/kg/min\\ Epinephrine \textless{}= 0.1 ug/kg/min\\ Dopamine \textgreater 5 ug/kg/min\end{tabular} & 6.0 - 11.9 {[}102 - 204{]} & \textless 50 & \begin{tabular}[c]{@{}c@{}}3.5 - 4.9\\ Urine output \textless 500\end{tabular} \\
4 & \begin{tabular}[c]{@{}c@{}}\textless 100 with respiratory\\ support\end{tabular} & \textless 6 & \begin{tabular}[c]{@{}c@{}}Norepinephrine \textgreater 0.1 ug/kg/min\\ Epinephrine \textgreater 0.1 ug/kg/min\\ Dopamine \textgreater 15 ug/kg/min\end{tabular} & \textgreater 12.0 {[}\textgreater 204{]} & \textless 20 & \begin{tabular}[c]{@{}c@{}}\textgreater 5.0\\ Urine output \textless 200\end{tabular} \\ \hline
\end{tabular}%
}
\end{table*}

\subsubsection{MIMIC-IV LoS labels}

For the remaining LoS prediction, we avoided the task definition by \citet{MIMICBenchmark} and formulated the task as time-to-event regression with an accelerated failure time objective. More specifically, we assume that covariates (EHR TS variables) would hasten or prolong the patient's discharge from ICU. The aim is to predict days until discharge at at each prediction point and we imposed right censoring at 30 days. Given that the utility of LoS prediction is mostly for ICU bed management, prediction over arbitrarily long horizon is not clinically useful due the presence of extreme outliers (\textbf{Table \ref{tab_rolling_stat}}). In our case, we consider 30 days to be a generous horizon as less than $5\%$ of the prediction points are censored.

\begin{algorithm}
\caption{Masked negative log-likelihood for Log-normal AFT}
\label{algorithm_aft_loss}
\begin{algorithmic}[1]
\REQUIRE $\mu,\log\sigma,y,\mathrm{event},\mathrm{mask}\in\mathbb{R}^{B\times T}$, \(\varepsilon>0\)
\STATE $\sigma \gets \exp(\log\sigma)$
\STATE $y \gets \max(y,\varepsilon)$ 
\STATE $\log y \gets \ln(y)$
\STATE $z \gets (\log y - \mu) / \sigma$
\STATE $\log f \gets -\log\big(y\cdot\sigma\sqrt{2\pi}\big) - \tfrac{1}{2}z^{2}$
\STATE $\log S \gets \log\big(1 - \Phi(z) + \varepsilon\big)$ 
\STATE $\mathrm{nll} \gets -\big(\mathrm{event}\cdot\log f + (1-\mathrm{event})\cdot\log S\big)$
\STATE $\mathrm{nll} \gets \mathrm{nll}\odot\mathrm{mask}$ 
\STATE $\mathrm{loss} \gets \dfrac{\sum \mathrm{nll}}{\sum \mathrm{mask}}$
\STATE \RETURN $\mathrm{loss}$
\end{algorithmic}
\end{algorithm}

The log-normal AFT objective for the linear and XGB baseline models were implemented by \citet{lifelines} and \citet{xgboost} respectively. For remaining LoS prediction, LLM4EHR and other deep learning baselines are fitted with a accelerated failure prediction head that predicts log-scale mean $\mu$ and standard deviation $\log\sigma$ of the expected time to discharge. Given the observed remaining LoS $y$, binary event indicator $event$, validity mask $m$ and batch size $B$, the negative log-likelihood is calculated via \textbf{Algorithm \ref{algorithm_aft_loss}}, which is derived from the implementation by \citet{weibullrnn}. The LoS prediction evaluation metrics for all baseline models were implemented by \citet{lifelines}.

\subsubsection{Summary EHR TS features}

\begin{table*}[ht]
\centering
\caption{Summary of EHR TS features for MIMIC-IV (A) and Physionet2012 (B), Count refers to the number of measurements for each variable in the dataset}
\label{tab_ehr_TS_summary}
\begin{tabular}{l l r r r r}
\toprule
\textbf{Variable} & \textbf{Unit} & \textbf{Count (A)} & \textbf{Median (A)} & \textbf{Count (B)} & \textbf{Median (B)} \\
\midrule
heart\_rate & bpm & 5546072 & 85.0 & 345998 & 86.0 \\
sbp & mmHg & 5195562 & 119.0 & 207385 & 118.0 \\
dbp & mmHg & 5194886 & 62.0 & 207362 & 58.0 \\
mbp & mmHg & 5213104 & 78.0 & 206267 & 78.0 \\
sbp\_ni & mmHg & 3254534 & 119.0 & 162383 & 118.0 \\
dbp\_ni & mmHg & 3253952 & 65.0 & 162167 & 57.0 \\
mbp\_ni & mmHg & 3266865 & 79.0 & 160089 & 76.0 \\
resp\_rate & breaths/min & 5452082 & 20.0 & 93142 & 19.0 \\
temperature & \textdegree C & 1698165 & 36.94 & 141273 & 37.1 \\
sao2 & \% & 5326804 & 97.0 & 15233 & 97.0 \\
glucose & mg/dL & 1216399 & 136.0 & 26115 & 127.0 \\
albumin & g/dL & 69185 & 3.0 & 4802 & 2.9 \\
bicarbonate & mmol/L & 450467 & 24.0 & 27169 & 23.0 \\
bun & mg/dL & 448356 & 24.0 & 27756 & 20.0 \\
creatinine & mg/dL & 449072 & 1.1 & 27895 & 1.0 \\
sodium & mEq/L & 477454 & 139.0 & 27227 & 139.0 \\
potassium & mEq/L & 475875 & 4.1 & 29041 & 4.1 \\
ph & pH Unit & 392576 & 7.38 & 46370 & 7.38 \\
lactate & mmol/L & 211644 & 1.8 & 16051 & 2.0 \\
fio2 & \%  & 283447 & 50.0 & 60770 & 50.0 \\
pao2 & mmHg & 392740 & 98.0 & 44346 & 120.0 \\
paco2 & mmHg & 392382 & 41.0 & 44404 & 39.0 \\
hematocrit & \% & 470665 & 28.9 & 36512 & 30.3 \\
alt & IU/L & 129481 & 37.0 & 6469 & 43.0 \\
alp & IU/L & 128635 & 99.0 & 6291 & 82.0 \\
ast & IU/L & 131084 & 52.0 & 6473 & 62.0 \\
bilirubin\_total & mg/dL & 129401 & 0.9 & 6511 & 0.9 \\
platelet & 10\textsuperscript{3}/\textmu L & 403151 & 179.0 & 28290 & 172.0 \\
wbc & 10\textsuperscript{3}/\textmu L & 396604 & 10.9 & 25891 & 11.4 \\
troponin\_t & ug/L & 40840 & 0.16 & 5101 & 0.30 \\
gcs & score (3--15) & 1593520 & 15.0 & 123112 & 13.0 \\
urineoutput & mL & 2955401 & 100.0 & 265848 & 70.0 \\
\bottomrule
\end{tabular}
\end{table*}

\textbf{Table \ref{tab_ehr_TS_summary}} summarises EHR TS features for MIMIC-IV and Physionet2012, with count of measurements and median values for each variable. The density of measurements is substantially higher in MIMIC-IV due to our MIMIC-IV dataset includes more patients and longer patient trajectories than Physionet2012. The EHR TS features for MIMIC-IV and Physionet2012 were synchronised to ICU stay hours following the curation process by \citet{mimicrepo}. Prior to analysis, physically implausible outliers were removed on a variable-by-variable basis. We imputed missing values by front filling within each episode, and filled any remaining gaps with the population mean. Continuous variables were z-standardised using observed mean and standard deviation before imputation.  

\end{document}